# Automatic Aggregation by Joint Modeling of Aspects and Values

**Christina Sauper**　　　　　　　　　　　　　　　　CSAUPER@CSAIL.MIT.EDU
**Regina Barzilay**　　　　　　　　　　　　　　　　REGINA@CSAIL.MIT.EDU
*Computer Science and Artificial Intelligence Laboratory*
*Massachusetts Institute of Technology*
*32 Vassar St.*
*Cambridge, MA 02139 USA*

## Abstract

We present a model for aggregation of product review snippets by joint aspect identification and sentiment analysis. Our model simultaneously identifies an underlying set of ratable aspects presented in the reviews of a product (e.g., *sushi* and *miso* for a Japanese restaurant) and determines the corresponding sentiment of each aspect. This approach directly enables discovery of highly-rated or inconsistent aspects of a product. Our generative model admits an efficient variational mean-field inference algorithm. It is also easily extensible, and we describe several modifications and their effects on model structure and inference. We test our model on two tasks, joint aspect identification and sentiment analysis on a set of Yelp reviews and aspect identification alone on a set of medical summaries. We evaluate the performance of the model on aspect identification, sentiment analysis, and per-word labeling accuracy. We demonstrate that our model outperforms applicable baselines by a considerable margin, yielding up to 32% relative error reduction on aspect identification and up to 20% relative error reduction on sentiment analysis.

## 1. Introduction

Online product reviews have become an increasingly valuable and influential source of information for consumers. The ability to explore a range of opinions allows consumers to both form a general opinion of a product and gather information about its positive and negative aspects (e.g., *packaging* or *battery life*). However, as more reviews are added over time, the problem of information overload gets progressively worse. For example, out of hundreds of reviews for a restaurant, most consumers will read only a handful before making a decision. In this work, our goal is to summarize a large number of reviews by discovering the most informational product aspects and their associated user sentiment.

To address this need, online retailers often use simple aggregation mechanisms to represent the spectrum of user sentiment. Many sites, such as Amazon, simply present a distribution over user-assigned star ratings, but this approach lacks any reasoning about *why* the products are given that rating. Some retailers use further breakdowns by specific predefined domain-specific aspects, such as *food*, *service*, and *atmosphere* for a restaurant. These breakdowns continue to assist in effective aggregation; however, because the aspects are predefined, they are generic to the particular domain and there is no further explanation of *why* one aspect was rated well or poorly. Instead, for truly informative aggregation, each product needs to be assigned a set of fine-grained aspects specifically tailored to that product.

The goal of our work is to provide a mechanism for effective unsupervised content aggregation able to discover specific, fine-grained aspects and associated values. Specifically, we represent each data set as a collection of *entities*; for instance, these can represent products in the domain of online reviews. We are interested in discovering fine-grained *aspects* of each entity (e.g., *sandwiches* or *dessert* for a restaurant). Additionally, we would like to recover a *value* associated with the aspect (e.g., sentiment for product reviews). A summary of the input and output can be found in Figure 1. Our input consists of short text snippets from multiple reviews for each of several products. In the restaurant domain, as in Figure 1, these are restaurants. We assume that each snippet is opinion-bearing and discusses one of the aspects which are relevant for that particular product. Our output consists of a set of dynamic (i.e., not pre-specified) aspects for each product, snippets labeled with the aspect which they discuss, and sentiment values for each snippet individually and each aspect as a whole. In Figure 1, the aspects identified for *Tasca Spanish Tapas* include *chicken*, *dessert*, and *drinks*, and the snippets are labeled with the aspects they describe and the correct polarity.

One way to approach this problem is to treat it as a multi-class classification problem. Given a set of predefined domain-specific aspects, it would be fairly straightforward for humans to identify which aspect a particular snippet describes. However, for our task of discovering fine-grained entity-specific aspects, there is no way to know a priori which aspects may be present across the entire data set or to provide training data for each; instead, we must select the aspects dynamically. Intuitively, one potential solution is to cluster the input snippets, grouping those which are lexically similar without prior knowledge of the aspects they represent. However, without some knowledge of which words represent the aspect for a given snippet, the clusters may not align to ones useful for cross-review analysis. Consider, for example, the two clusters of restaurant review snippets shown in Figure 2. While both clusters share many words among their members, only the first describes a coherent aspect cluster, namely the *drinks* aspect. The snippets of the second cluster do not discuss a single product aspect, but instead share expressions of sentiment.

To successfully navigate this challenge, we must distinguish between words which indicate aspect, words which indicate sentiment, and extraneous words which do neither. For both aspect identification and sentiment analysis, it is crucial to know which words within a snippet are relevant for the task. Distinguishing them is not straightforward, however. Some work in sentiment analysis relies on a predefined lexicon or WordNet to provide some hints, but there is no way to anticipate every possible expression of aspect or sentiment, especially in user-generated data (e.g., use of slang such as "deeeeeee-lish" for "delicious"). In lieu of an explicit lexicon, we can attempt to use other information as a proxy, such as part of speech; for example, aspect words are likely to be nouns, while value words are more likely to be adjectives. However, as we show later in this paper, this additional information is again not sufficient for the tasks at hand.

Instead, we propose an approach to analyze a collection of product review snippets and jointly induce a set of learned aspects, each with a respective value (e.g., sentiment). We capture this idea using a generative Bayesian topic model where the set of aspects and any corresponding values are represented as hidden variables. The model takes a collection of snippets as input and explains how the observed text arises from the latent variables, thereby connecting text fragments with the corresponding aspects and values.

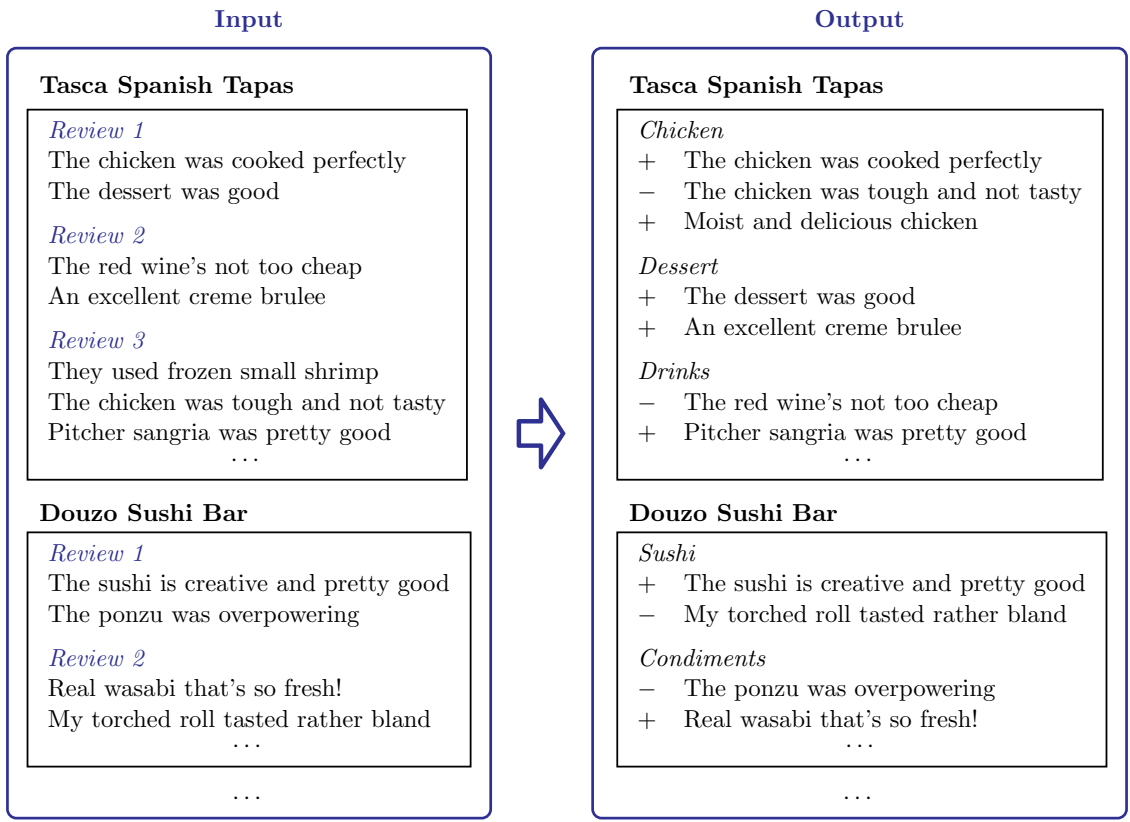

Figure 1: An example of the desired input and output of our system in the restaurant domain. The input consists of a collection of review snippets for several restaurants. The output is an aggregation of snippets by aspect (e.g., *chicken* and *dessert*) along with an associated sentiment for each snippet. Note that the input data is completely unannotated; the only information given is which snippets describe the same restaurant.

Specifically, we begin by defining sets of sentiment word distributions and aspect word distributions. Because we expect the types of sentiment words to be consistent across all products (e.g., any product may be labeled as "great" or "terrible"), we allow the positive and negative sentiment word distributions to be shared across all products. On the other hand, in the case of restaurant reviews and similar domains, aspect words are expected to be quite distinct between products. Therefore, we assign each product its own set of aspect word distributions. In addition to these word distributions, our model takes into account several other factors. First, we model the idea that each particular aspect of a product has some underlying quality; that is, if there are already 19 snippets praising a particular aspect, it's likely that the 20th snippet will be positive as well. Second, we account for common patterns in language using a transition distribution between types of words. For example, it is very common to see the pattern "`Value Aspect`," such as in phrases like "great pasta." Third, we model the distributions over parts of speech for each type of distribution. This covers the intuition that aspect words are frequently nouns, whereas value words are often adjectives. We describe each of these factors and our model as a whole in detail in Section 4.

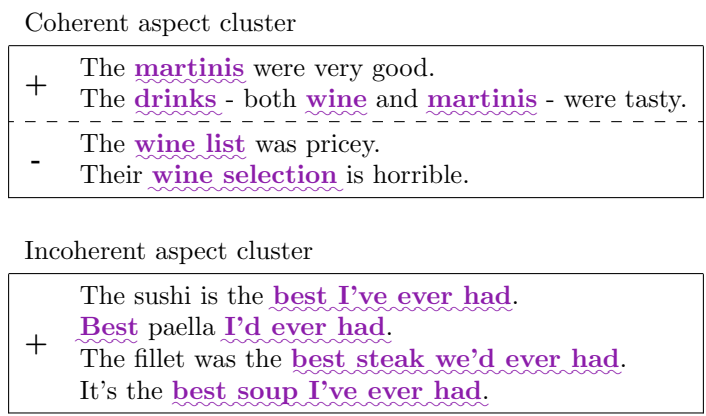

Figure 2: Example clusters of restaurant review snippets generated by a lexical clustering algorithm; words relevant to clustering are highlighted. The first cluster represents a coherent *aspect* of the underlying product, namely the *drinks* aspect. The latter cluster simply shares a common sentiment expression and does not represent snippets discussing the same product aspect. In this work, we aim to produce the first type of aspect cluster along with the corresponding values.

This formulation provides several advantages: First, the model does not require a set of predefined aspects. Instead, it is capable of assigning latent variables to discover the appropriate aspects based on the data. Second, the joint analysis of aspect and value allows us to leverage several pieces of information to determine which words are relevant for aspect identification and which should be used for sentiment analysis, including part of speech and global or entity-specific distributions of words. Third, the Bayesian model admits an efficient mean-field variational inference procedure which can be parallelized and run quickly on even large numbers of entities and snippets.

We evaluate our approach on the domain of restaurant reviews. Specifically, we use a set of snippets automatically extracted from restaurant reviews on Yelp. This collection consists of an average of 42 snippets for each of 328 restaurants in the Boston area, representing a wide spectrum of opinions about several aspects of each restaurant. We demonstrate that our model can accurately identify clusters of review fragments that describe the same aspect, yielding 32.5% relative error reduction (9.9 absolute $F_1$) over a standalone clustering baseline. We also show that the model can effectively identify snippet sentiment, with a 19.7% relative error reduction (4.3% absolute accuracy) over applicable baselines. Finally, we test the model's ability to correctly label aspect and sentiment words, discovering that the aspect identification has high-precision, while the sentiment identification has high-recall.

Additionally, we apply a slimmed-down version of our model which focuses exclusively on aspect identification to a set of lab- and exam-related snippets from medical summaries provided by the Pediatric Environmental Health Clinic (PEHC) at Children's Hospital Boston. These summaries represent concise overviews of the patient information at a par-ticular visit, as relayed from the PEHC doctor to the child's referring physician. Our model achieves 7.4% (0.7 absolute $F_1$) over the standalone clustering baseline.

The remainder of this paper is structured as follows. Section 2 compares our work with previous work on both aspect identification and sentiment analysis. Section 3 describes our specific problem formulation and task setup more concretely. Section 4 presents the details of our full model and various model extensions, and Section 5 describes the inference procedure and the necessary adjustments for each extension. The details of both data sets, the experimental formulation, and results are presented in Section 6. We summarize our findings and consider directions for future work in Section 7. The code and data used in this paper are available online at `http://groups.csail.mit.edu/rbg/code/review-aggregation`.

## 2. Related Work

Our work falls into the area of multi-aspect sentiment analysis. In this section, we first describe approaches toward document-level and sentence-level sentiment analysis (Section 2.1), which provide the foundation for future work, including our own. Then, we describe three common directions of multi-aspect sentiment analysis; specifically, those which use data-mining or fixed-aspect analysis (Section 2.2.1), those which incorporate sentiment analysis with multi-document summarization (Section 2.2.2), and finally, those focused on topic modeling with additional sentiment components (Section 2.2.3).

### 2.1 Single-Aspect Sentiment Analysis

Early sentiment analysis focused primarily on identification of coarse document-level sentiment (Pang, Lee, & Vaithyanathan, 2002; Turney, 2002; Pang & Lee, 2008). Specifically, these approaches attempted to determine the overall polarity of documents. These approaches included both rule-based and machine learning approaches: Turney (2002) used a rule-based method to extract potentially sentiment-bearing phrases and then compared them to the sentiment of known-polarity words, while Pang et al. (2002) used discriminative methods with features such as unigrams, bigrams, part-of-speech tags, and word position information.

While document-level sentiment analysis can give us the overall view of an opinion, looking at individual sentences within the document yields a more fine-grained analysis. The work in sentence-level sentiment analysis focuses on first identifying sentiment-bearing sentences and then determining their polarity (Yu & Hatzivassiloglou, 2003; Dave, Lawrence, & Pennock, 2003; Kim & Hovy, 2005, 2006; Pang & Lee, 2008). Both identification of sentiment-bearing sentences and polarity analysis can be performed through supervised classifiers (Yu & Hatzivassiloglou, 2003; Dave et al., 2003) or similarity to known text (Yu & Hatzivassiloglou, 2003; Kim & Hovy, 2005), through measures based on distributional similarity or by using WordNet relationships.

By recognizing connections between parts of a document, sentiment analysis can be further improved (Pang & Lee, 2004; McDonald, Hannan, Neylon, Wells, & Reynar, 2007; Pang & Lee, 2008). Pang and Lee (2004) leverage the relationship between sentences to improve document-level sentiment analysis. Specifically, they utilize both the subjectivity of individual sentences and information about the strength of connection between sentences in a min cut formulation to provide better sentiment-focused summaries of text. McDonald et al. (2007) examine a different connection, instead constructing a hierarchical model of sentiment between sentences and documents. Their model uses complete labeling on a subset of data to learn a generalized set of parameters which improve classification accuracy at both document-level and sentence-level.

While none of the above approaches attempt to identify aspects or analyze sentiment in an aspect-based fashion, the intuitions provide key insight into the approaches we take in our work. For example, the importance of distinguishing opinion sentences follows our own intuition about the necessity of identifying sentiment-bearing words within a snippet.

## 2.2 Aspect-Based Sentiment Analysis

Following the work in single-aspect document-level and sentence-level sentiment analysis came the intuition of modeling aspect-based (also called "feature-based") sentiment for review analysis. We can divide these approaches roughly into three types of systems based on their techniques: systems which use fixed-aspect approaches or data-mining techniques for aspect selection or sentiment analysis, systems which adapt techniques from multi-document summarization, and systems which jointly model aspect and sentiment with probabilistic topic models. Here, we examine each avenue of work with relevant examples and contrast them with our own work.

### 2.2.1 Data-Mining and Fixed-Aspect Techniques for Sentiment Analysis

One set of approaches toward aspect-based sentiment analysis follow the traditional techniques of data mining (Hu & Liu, 2004; Liu, Hu, & Cheng, 2005; Popescu, Nguyen, & Etzioni, 2005). These systems may operate on full documents or on snippets, and they generally require rule-based templates or additional resources such as WordNet both to identify aspects and to determine sentiment polarity. Another approach is to fix a predetermined relevant set of aspects, then focus on learning the optimal opinion assignment for these aspects (Snyder & Barzilay, 2007). Below, we summarize each approach and compare and contrast them to our work.

One set of work relies on a combination of association mining and rule-based extraction of nouns and noun phrases for aspect identification. Hu and Liu (2004) and Liu et al. (2005) developed a three-step system: First, initial aspects are selected by an association miner and pruned by a series of rules. Second, related opinions for each aspect are identified in a rule-based fashion using word positions, and their polarity is determined by WordNet search based on a set of seed words. Third, additional aspects are identified in a similar fashion based on position of the selected polarity words. In each of these steps, part-of-speech information provides a key role in the extraction rules. In the later work, there is an additional component to identify *implicit* aspects in a deterministic fashion; e.g., *heavy* maps deterministically to <WEIGHT> (Liu et al., 2005). While their task is similar to ours and we utilize part-of-speech information as an important feature as well, we additionally leverage other distributional information to identify aspects and sentiment. Furthermore, we avoid the reliance on WordNet and predefined rule mappings in order to preserve the generality of the system. Instead, our joint modeling allows us to recover these relationships without the need for additional information.

Other approaches also rely on WordNet relationships to identify not only sentiment polarity, but also aspects, using the *parts* and *properties* of a particular product class. Popescu et al. (2005) first use these relations to generate the set of aspects for a given product class (e.g., *camera*). Following that, they apply relaxation labeling for sentiment analysis. This procedure gradually expands sentiment from individual words to aspects to sentences, similar to the *Cascade* pattern mentioned in the work of McDonald et al. (2007). Like the system of Liu et al. (2005), their system requires a set of manual rules and several outside resources. While our model does require a few seed words, it does not require any manual rules or additional resources due to its joint formulation.

A separate direction of work relies on predefined aspects while focusing on improvement of sentiment analysis prediction. Snyder and Barzilay (2007) define a set of aspects specific to the restaurant domain. Specifically they define an individual rating model for each aspect, plus an overall agreement model which attempts to determine whether the resulting ratings should all agree or disagree. These models are jointly trained in a supervised fashion using an extension of the PRanking algorithm (Crammer & Singer, 2001) to find the best overall star rating for each aspect. Our problem formulation differs significantly from their work in several dimensions: First, we desire a more refined analysis using fine-grained aspects instead of coarse predefined features. Second, we would like to use as little supervised training data as possible, rather than the supervised training required for the PRanking algorithm.

In our work, we attempt to capture the intuitions of these approaches while reducing the need for outside resources and rule-based components. For example, rather than supplying rule-based patterns for extraction of aspect and sentiment, we instead leverage distributional patterns across the corpus to infer the relationships between words of different types. Likewise, rather than relying on WordNet relationships such as synonymy, antonymy, hyponymy, or hypernymy (Hu & Liu, 2004; Liu et al., 2005; Popescu et al., 2005), we bootstrap our model from a small set of seed words.

### 2.2.2 Multi-Document Summarization and its Application to Sentiment Analysis

Multi-document summarization techniques generally look for repetition across documents to signal important information (Radev & McKeown, 1998; Barzilay, McKeown, & Elhadad, 1999; Radev, Jing, & Budzikowska, 2000; Mani, 2001). For aspect-based sentiment analysis, work has focused on augmenting these techniques with additional components for sentiment analysis (Seki, Eguchi, Kanodo, & Aono, 2005, 2006; Carenini, Ng, & Pauls, 2006; Kim & Zhai, 2009). In general, the end goal of these approaches is the task of forming coherent text summaries using either text extraction or natural language generation. Unlike our work, many of these approaches do not explicitly identify aspects; instead, they are extracted through repeated information. Additionally, our model explicitly looks at the connection between content and sentiment, rather than treating it as a secondary computation after information has been selected.

One technique for incorporating sentiment analysis follows previous work on identification of opinion-bearing sentences. Seki et al. (2005, 2006) present DUC summarization systems designed to create opinion-focused summaries of task topics.[1] In their system, they employ a subjectivity component using a supervised SVM with lexical features, similar to those in the work of Yu and Hatzivassiloglou (2003) and Dave et al. (2003). This component is used to identify subjective sentences and, in the work of Seki et al. (2006), their polarity, both in the task and in the sentences selected for the response summary. However, like previous work and unlike our task, there is no aspect-based analysis in their summarization task. It is also fully supervised, relying on a hand-annotated set of about 10,000 sentences to train the SVM.

Another line of work focuses on augmenting the summarization system with aspect selection similar to the data-mining approaches of Hu and Liu (2004), rather than using single-aspect analysis. Carenini, Ng, and Zwart (2005) and Carenini et al. (2006) augment the previous aspect selection with a user-defined hierarchical organization over aspects; e.g., *digital zoom* is part of the *lens*. Polarity of each aspect is assumed to be given by previous work. These aspects are then incorporated into existing summarization systems – MEAD* sentence extraction (Radev et al., 2000) or SEA natural language generation (Carenini & Moore, 2006) – to form final summaries. Like the work of Seki et al. (2005, 2006), this work does not create new techniques for aspect identification or sentiment analysis; instead, they focus on the process of integrating these sources of information with summarization systems. While the aspects produced are comparable across reviews for a particular product, the highly-supervised nature means that this approach is not feasible for a large set of products such as our corpus of reviews from many types of restaurants. Instead, we must be able to dynamically identify relevant aspects.

A final line of related work relies on the traditional summarization technique of identifying contrastive or contradictory sentences. Kim and Zhai (2009) focus on generating contrastive summaries by identifying pairs of sentences which express differing opinions on a particular product feature. To do this, they define metrics of *representativeness* (coverage of opinions) and *contrastiveness* (alignment quality) using both semantic similarity with WordNet matches and word overlap. In comparison to our work, this approach follows an orthogonal goal, as we try to find the most defining aspects instead of the most contradictory ones. Additionally, while the selected pairs hint at disagreements in rating, there is no identification of how many people agree with each side or the overall rating of a particular aspect. In our work, we aim to produce both a concrete set of aspects and the user sentiment for each, whether it is unanimous or shows disagreement.

Overall, while these methods are designed to produce output summaries which focus on subjective information, they are not specifically targeted for aspect-based analysis. Instead, aspects are identified in a supervised fashion (Carenini et al., 2005, 2006) or are not defined at all (Seki et al., 2005, 2006; Kim & Zhai, 2009). In our work, it is crucial that we have dynamically-selected aspects because it is not feasible to preselect aspects in a supervised fashion.

### 2.2.3 Probabilistic Topic Modeling for Sentiment Analysis

The work closest to our own in the direction of aspect-based analysis focuses on the use of probabilistic topic modeling techniques for identification of aspects. These may be aggre-gated without specific sentiment polarity (Lu & Zhai, 2008) or combined with additional sentiment modeling either jointly (Mei, Ling, Wondra, Su, & Zhai, 2007; Blei & McAuliffe, 2008; Titov & McDonald, 2008a) or as a separate post-processing step (Titov & McDonald, 2008b). Like our work, these approaches share the intuition that aspects may be represented as topics.

---

1. For task examples, see the work of Dang (2005, 2006).

Several approaches focus on extraction of topics and sentiment from blog articles. In one approach, they are used as expert articles for aspect extraction in combination with a larger corpus of user reviews. Lu and Zhai (2008) introduce a model with semi-supervised probabilistic latent semantic analysis (PLSA) which identifies sentiment-bearing aspects through segmentation of an expert review. Then, the model extracts compatible supporting and supplementary text for each aspect from the set of user reviews. Aspect selection is constrained as in the rule-based approaches; specifically, aspect words are required to be nouns. Our work differs from their work significantly. While we share a common goal of identifying and aggregating opinion-bearing aspects, we additionally desire to identify the polarity of opinions, a task not addressed in their work. In addition, obtaining aspects from an expert review is unnecessarily constraining; in practice, while expert reviewers may mention some key aspects, they will not mention every aspect. It is crucial to discover aspects based on the entire set of articles.

There is work in the direction of aspect identification from blog posts. For example, Mei et al. (2007) use a variation on latent Dirichlet allocation (LDA) similar to our own to explicitly model both topics and sentiment, then use a hidden Markov model to discover sentiment dynamics across topic life cycles. A general sentiment polarity distribution is computed by combining distributions from several separate labeled data sets (e.g., movies, cities, etc.). However, in their work, sentiment is measured at the document-level, rather than topic-level. Additionally, the topics discovered by their model are very broad; for example, when processing the query "The Da Vinci Code", returned topics may be labeled as *book*, *movie*, and *religion*, rather than the fine-grained aspects we desire in our model, such as those representing major characters or events. Our model expands on their work by discovering very fine-grained aspects and associating particular sentiment with each individual aspect. In addition, by tying sentiment to aspects, we are able to identify sentiment-bearing words and their associated polarities without the additional annotation required to train an external sentiment model.

Sentiment may also be combined with LDA using additional latent variables for each document in order to predict document-level sentiment. Blei and McAuliffe (2008) propose a form of supervised LDA (sLDA) which incorporates an additional response variable, which can be used to represent sentiment such as the star rating of a movie. They can then jointly model the documents and responses in order to find the latent topics which best predict the response variables for future unlabeled documents. This work is significantly different from our work, as it is supervised and does not predict in a multi-aspect framework.

Building on these approaches comes work in fine-grained aspect identification with sentiment analysis. Titov and McDonald (2008a, 2008b) introduce a multi-grain unsupervised topic model, specifically built as an extension to LDA. This technique yields a mixture of global and local topics. Word distributions for all topics (both global and local) are drawn at the global level, however; unlike our model. The consequence of this is that topics are very easy to compare across all products in the corpus; however, the topics are more gen-eral and less dynamic than we hope to achieve because they must be shared among every product. One consequence of defining global topics is difficulty in finding relevant topics for every product when there is little overlap. For example, in the case of restaurant reviews, Italian restaurants should have a completely different set of aspects than Indian restaurants. Of course, if these factors were known, it would be possible to run the algorithm separately on each subset of restaurants, but these distinctions are not immediately clear a priori. Increasing the number of topics could assist in recovering additional aspects; however because the aspects are still global, it will still be difficult to identify restaurant-specific aspects.

For sentiment analysis, the PRanking algorithm of Snyder and Barzilay (2007) is incorporated in two ways: First, the PRanking algorithm is trained in a pipeline fashion after all topics are generated (Titov & McDonald, 2008b); later, it is incorporated into the model during inference in a joint formulation (Titov & McDonald, 2008a). However, in both cases, as in the original algorithm, the set of aspects is fixed – each of the aspects corresponds to a fixed set of of topics found by the model. Additionally, the learning problem is supervised. Because of the fixed aspects, necessary additional supervision, and global topic distribution, this model formulation is not sufficient for our problem domain, which requires very fine-grained aspects.

All of these approaches have structural similarity to the work we present, as they are variations on LDA. None, however, has the same intent as our model. Mei et al. (2007) model aspect and sentiment jointly; however their aspects are very vague, and they treat sentiment at the document level rather than the aspect level. Likewise, Titov and McDonald (2008b, 2008a) model "fine-grained" aspects, but they are still coarser than the aspects we require, even if we were to increase the number of aspects, as their distributions are shared globally. Finally, Lu and Zhai (2008), Blei and McAuliffe (2008), and Titov and McDonald (2008b, 2008a) require supervised annotation or a supervised expert review that we do not have. We attempt to solve each of these issues with our joint formulation in order to proceed with minimal supervision and discover truly fine-grained aspects.

## 3. Problem Formulation

Before explaining the model details, we describe the random variables and abstractions of our model, as well as some intuitions and assumptions.[2] A visual explanation of model components is shown in Figure 3. We present complete details and the generative story in Section 4.

### 3.1 Model Components

Our model is composed of five component types: entities, snippets, aspects, values, and word topics. Here, we describe each type and provide examples.

---

2. Here, we explain our complete model with value selection for sentiment in the restaurant domain. For the simplified case in the medical domain where we would like to use only aspects, we may simply ignore the value-related components of the model.

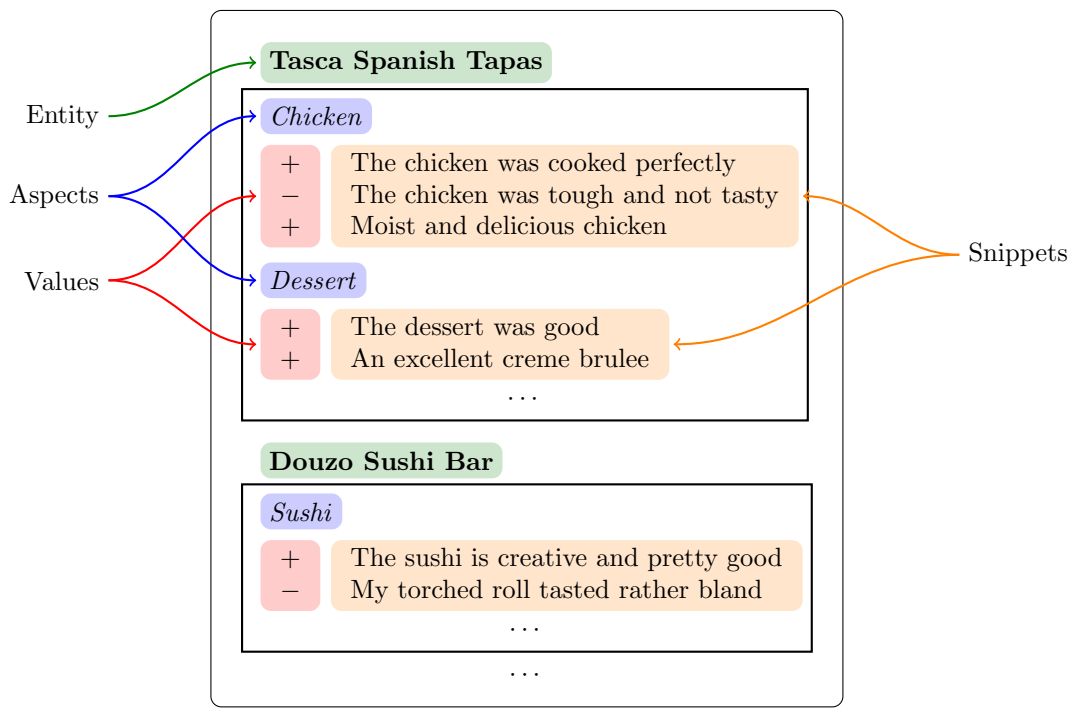

Figure 3: Labeled model components from the example in Figure 1. Note that aspects are never given explicit labels, and the ones shown here are presented purely for ease of understanding; aspects exist simply as groups of snippets which share a common subject. Also, word topics are not pictured here; a word topic (Aspect, Value, or Background) is assigned to each word in each snippet. These model components are described at high level in Section 3.1 and in depth in Section 4.

### 3.1.1 Entity

An entity represents a single object which is described in the review. In the restaurant domain, these represent individual restaurants, such as *Tasca Spanish Tapas*, *Douzo Sushi Bar*, and *Outback Steakhouse*.

### 3.1.2 Snippet

A snippet is a user-generated short sequence of words describing an entity. These snippets can be provided by the user as is (for example, in a "quick reaction" box) or extracted from complete reviews through a phrase extraction system such as the one from Sauper, Haghighi, and Barzilay (2010). We assume that each snippet contains at most one single aspect (e.g., *pizza*) and one single value type (e.g., *positive*). In the restaurant domain, this corresponds to giving an opinion about one particular dish or category of dishes. Examples from the restaurant domain include *"Their pasta dishes are perfection itself"*, *"they had fantastic drinks"*, and *"the lasagna rustica was cooked perfectly"*.

### 3.1.3 Aspect

An aspect corresponds to one of several properties of an entity. In the restaurant domain where entities represent restaurants, aspects may correspond to individual dishes or categories of dishes, such as *pizza* or *alcoholic drinks*. For this domain, each entity has its own unique set of aspects. This allows us to model aspects at the appropriate granularity. For example, an Italian restaurant may have a *dessert* aspect which pertains to information about a variety of cakes, pies, and gelato. However, most of a bakery's menu would fall under that same *dessert* aspect. Instead, to present a useful aspect-based summary, it would require separate aspects for each of *cakes*, *pies*, and so on. Because aspects are entity-specific rather than shared, there are no ties between restaurants which have aspects in common (e.g., most sushi restaurants will have a *sashimi* aspect); we consider this a point for potential future work. Note that it is still possible to compare aspects across entities (e.g., to find the best restaurant for a *burger*) by comparing their respective word distributions.

### 3.1.4 Value

Values represent the information associated with an aspect. In the review domain, the two value types represent positive and negative sentiment respectively. In general, it is possible to use value to represent other distinctions; for example, in a domain where some aspects are associated with a numeric value and others are associated with a text description, each of these can be set as a value type. The intended distinctions may be encouraged by the use of seed words (see Section 3.2), or they may be left unspecified for the model to assign whatever it finds to best fit the data. The number of value types must be prespecified; however, it is possible to use either very few or very many types.

### 3.1.5 Word Topic

While the words in each snippet are observed, each word is associated with an underlying latent topic. The possible latent topics correspond to aspect, value, and a background topic. For example, in the review domain, the latent topic of words *great* or *terrible* would be `Value`, of words which represent entity aspects such as *pizza* would be `Aspect`, and of stop words like *is* or of in-domain white noise like *food* would be `Background`.

## 3.2 Problem Setup

In this work, we assume that the snippet words are always observed, and the correlation between snippets and entities is known (i.e., we know which entity a given snippet describes). In addition, we assume part of speech tags for each word in each snippet. As a final source of supervision, we may optionally include small sets of seed words for a lexical distribution, in order to bias the distribution toward the intended meaning. For example, in the sentiment case, we can add seed words in order to bias one value distribution toward positive and one toward negative. Seed words are certainly not required; they are simply a tool to constrain the model's use of distributions to fit any prior expectations.

Note that in this formulation, the relevant aspects for each restaurant are **not** observed; instead, they are represented by lexical distributions which are induced at inference time. In the system output, aspects are represented as unlabeled clusters over snippets.[3] Given this formulation, the goal of this work is then to induce the latent aspect and value underlying each snippet.

## 4. Model

Our model has a generative formulation over all snippets in the corpus. In this section, we first describe in detail the general formulation and notation of the model, then discuss novel changes and enhancements for particular corpora types. Inference for this model will be discussed in Section 5. As mentioned previously, we will describe the complete model including aspect values.

### 4.1 General Formulation

For this model, we assume a collection of all snippet words for all entities, $\mathbf{s}$. We use $s^{i,j,w}$ to denote the $w$th word of the $j$th snippet of the $i$th entity. We also assume a fixed vocabulary of words $W$.

We present a summary of notation in Table 1, a concise summary of the model in Figure 4, and a model diagram in Figure 5. There are three levels in the model design: global distributions common to all snippets for all entities in the collection, entity-level distributions common to all snippets describing a single entity, and snippet- and word-level random variables. Here, we describe each in turn.

#### 4.1.1 Global Distributions

At the global level, we draw a set of distributions common to all entities in the corpus. These include everything shared across a domain, such as the background stop-word distribution, value types, and word topic transitions.

**Background Distribution**  A global background word distribution $\theta_B$ is drawn to represent stop-words and in-domain white noise (e.g., "food" becomes white noise in a corpus of restaurant reviews). This distribution is drawn from a symmetric Dirichlet with concentration parameter $\lambda_B$; in our experiments, this is set to 0.2.

**Value Distributions**  A value word distribution $\theta_V^v$ is drawn for each value type $v$. For example, in a review domain with positive and negative sentiment types, there will be a distribution over words for the positive type and one for the negative type. Seed words $W_{seed_v}$ are given additional probability mass on the value priors for type $v$; specifically, a non-seed word receives $\epsilon$ hyperparameter, while a seed word receives $\epsilon + \lambda_V$; in our experiments, this is set to 0.15.

**Transition Distribution**  A transition distribution $\Lambda$ is drawn to represent the transition probabilities of underlying word topics. For example, it may be very likely to have a `Value Aspect` transition in a review domain, which fits phrases like "great pizza." In our experiments, this distribution is given a slight prior bias toward more helpful transitions; for

---

| Data Set | |
|---|---|
| $\mathbf{s}$ | Collection of all snippet words from all entities |
| $s^{i,j,w}$ | $w$th word of $j$th snippet of $i$th entity |
| $t^{i,j,w}$ * | Part-of-speech tag corresponding to $s^{i,j,w}$ |
| $W$ | Fixed vocabulary |
| $W_{seed_v}$ | Seed words for value type $v$ |

| Lexical Distributions | |
|---|---|
| $\theta_B$ | Background word distribution |
| $\theta_A^{i,a}$ ($\theta_A^a$ *) | Aspect word distribution for aspect $a$ of entity $i$ |
| $\theta_V^v$ | Value word distribution for type $v$ |
| $\theta_I$ * | Ignored words distribution |

| Other Distributions | |
|---|---|
| $\Lambda$ | Transition distribution over word topics |
| $\phi^{i,a}$ ($\phi^a$ *) | Aspect-value multinomial for aspect $a$ of entity $i$ |
| $\psi^i$ ($\psi$ *) | Aspect multinomial for entity $i$ |
| $\eta$ * | Part-of-speech tag distribution |

| Latent Variables | |
|---|---|
| $Z_A^{i,j}$ | Aspect selected for $s^{i,j}$ |
| $Z_V^{i,j}$ | Value type selected for $s^{i,j}$ |
| $Z_W^{i,j,w}$ | Word topic ($A$, $V$, $B$, $I$ *) selected for $s^{i,j,w}$ |

| Other Notation | |
|---|---|
| $K$ | Number of aspects $a$ |
| $A$ | Indicator corresponding to aspect word |
| $V$ | Indicator corresponding to value word |
| $B$ | Indicator corresponding to background word |
| $I$ * | Indicator corresponding to ignored word |

Table 1: Notation used in this paper. Items marked with * relate to extensions mentioned in Section 4.2.

example, encouraging sticky behavior by providing a small boost to self-transitions. This bias is easily overridden by data; however, it provides a useful starting point.

### 4.1.2 Entity-Specific Distributions

There are naturally variations in the aspects which snippets describe and how many snippets describe each aspect. For example, a mobile device popular for long battery life will likely have more snippets describing the battery than a device which is known for its large screen. Some domains may have enormous variation in aspect vocabulary; for example, in restaurant reviews, two restaurants may not serve any of the same food items to compare. To account

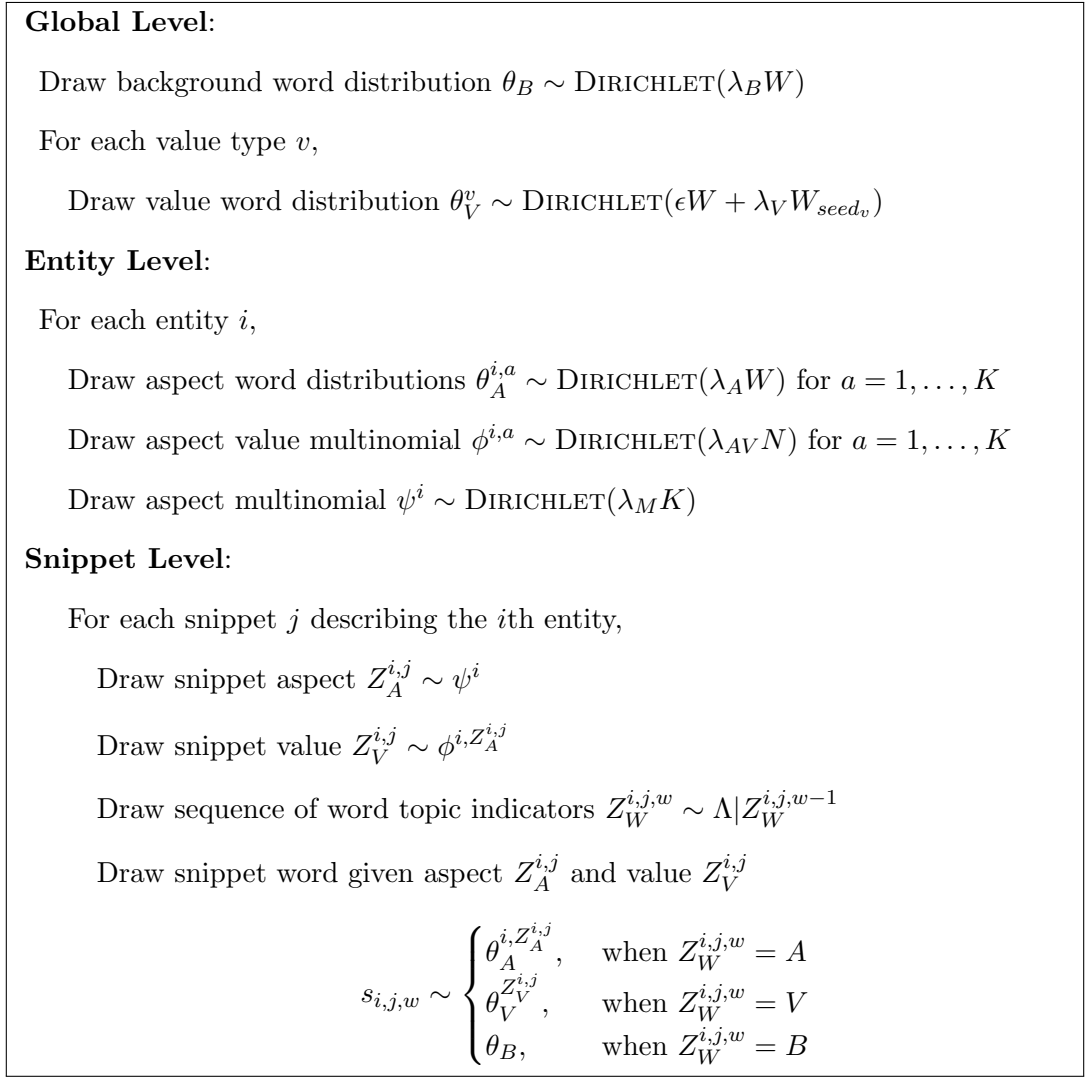

**Global Level**:

  Draw background word distribution $\theta_B \sim \textsc{Dirichlet}(\lambda_B W)$

  For each value type $v$,

    Draw value word distribution $\theta_V^v \sim \textsc{Dirichlet}(\epsilon W + \lambda_V W_{seed_v})$

**Entity Level**:

  For each entity $i$,

    Draw aspect word distributions $\theta_A^{i,a} \sim \textsc{Dirichlet}(\lambda_A W)$ for $a = 1, \ldots, K$

    Draw aspect value multinomial $\phi^{i,a} \sim \textsc{Dirichlet}(\lambda_{AV} N)$ for $a = 1, \ldots, K$

    Draw aspect multinomial $\psi^i \sim \textsc{Dirichlet}(\lambda_M K)$

**Snippet Level**:

  For each snippet $j$ describing the $i$th entity,

    Draw snippet aspect $Z_A^{i,j} \sim \psi^i$

    Draw snippet value $Z_V^{i,j} \sim \phi^{i,Z_A^{i,j}}$

    Draw sequence of word topic indicators $Z_W^{i,j,w} \sim \Lambda | Z_W^{i,j,w-1}$

    Draw snippet word given aspect $Z_A^{i,j}$ and value $Z_V^{i,j}$

$$s_{i,j,w} \sim \begin{cases} \theta_A^{i,Z_A^{i,j}}, & \text{when } Z_W^{i,j,w} = A \\ \theta_V^{Z_V^{i,j}}, & \text{when } Z_W^{i,j,w} = V \\ \theta_B, & \text{when } Z_W^{i,j,w} = B \end{cases}$$

Figure 4: A summary of our generative model presented in Section 4.1. We use $\textsc{Dirichlet}(\lambda W)$ to denote a finite Dirichlet prior where the hyper-parameter counts are a scalar times the unit vector of vocabulary items. For the global value word distribution, the prior hyper-parameter counts are $\epsilon$ for all vocabulary items and $\lambda_V$ for $W_{seed_v}$, the vector of vocabulary items in the set of seed words for value $v$.

for these variations, we define a set of entity-specific distributions which generate both aspect vocabulary and popularity, as well as a distribution over value types for each aspect.

**Aspect Distributions** An aspect word distribution $\theta_A^{i,a}$ is drawn for each aspect $a$. Each of these represents the distribution over unigrams for a particular aspect. For example, in the domain of restaurant reviews, aspects may correspond to menu items such as *pizza*, while in reviews for cell phones, they may correspond to details such as *battery life*. Each aspect word distribution is drawn from a symmetric Dirichlet prior with hyperparameter $\lambda_A$; in our experiments, this is set to 0.075.

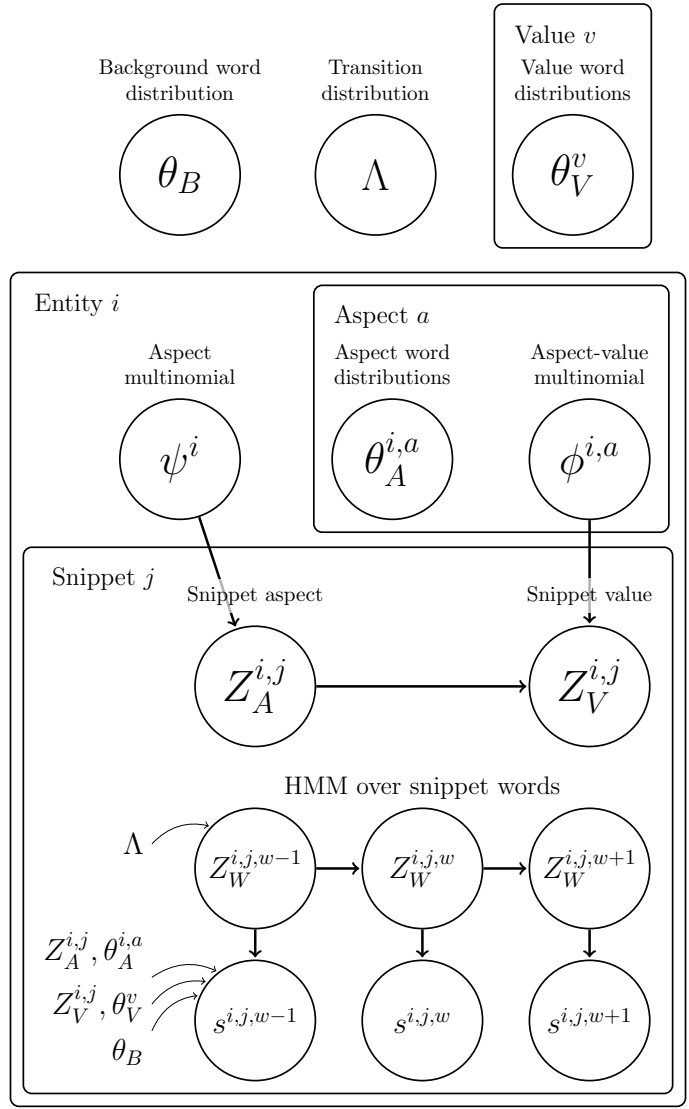

Figure 5: A graphical description of the model presented in Section 4.1. A written description of the generative process is located in Figure 4. Curved arrows indicate additional links which are present in the model but not drawn for readability.

**Aspect-Value Multinomials** Aspect-value multinomials $\phi^{i,a}$ determine the likelihood of each value type $v$ for the corresponding aspect $a$. For example, if value types represent positive and negative sentiment, this corresponds to agreement of sentiment across snippets. Likewise, if value types represent formatting such as integers, decimals, and text, each aspect generally prefers the same type of value. These multinomials are drawn from a symmetric Dirichlet prior using hyperparameter $\lambda_{AV}$; in our experiments, this is set to 1.0.

**Aspect Multinomial** The aspect multinomial $\psi^i$ controls the likelihood of each aspect being discussed in a given snippet. This encodes the intuition that certain aspects are more likely to be discussed than others for a given entity. For example, if a particular Italian restaurant is famous for their pizza, it is likely that the *pizza* aspect will be frequently discussed in reviews, while the *drinks* aspect may be mentioned only occasionally. The aspect multinomial will encode this as a higher likelihood for choosing *pizza* as a snippet aspect than *drinks*. This multinomial is drawn from a symmetric Dirichlet distribution with hyperparameter $\lambda_M$; in our experiments, this is set to 1.0.

### 4.1.3 Snippet- and Word-Specific Random Variables

Using the distributions described above, we can now draw random variables for each snippet to determine the aspect and value type which will be described, as well as the sequence of underlying word topics and words.

**Aspect** A single aspect $Z_A^{i,j}$ which this snippet will describe is drawn from the aspect multinomial $\psi^i$. All aspect words in the snippet (e.g., *pizza* in a corpus of restaurant reviews) will be drawn from the corresponding aspect word distribution $\theta_A^{i,Z_A^{i,j}}$.

**Value Type** A single value type $Z_V^{i,j}$ is drawn conditioned on the selected aspect from the corresponding aspect-value multinomial $\phi^{i,Z_A^{i,j}}$. All value words in the snippet (e.g., "great" in the review domain) will be drawn from the corresponding value word distribution $\theta_V^{Z_V^{i,j}}$.

**Word Topic Indicators** A sequence of word topic indicators $Z_W^{i,j,1}, \ldots, Z_W^{i,j,m}$ is generated using a first-order Markov model parameterized by the transition matrix $\Lambda$. These indicators determine which unigram distribution generates each word in the snippet. For example, if $Z_W^{i,j,w} = B$, the $w$th word of this snippet is generated from the background word distribution $\theta_B$.

## 4.2 Model Extensions

There are a few optional components of the model which may improve performance for some cases. We briefly list them here, then present the necessary modifications to the model in detail for each case. Modifications to the inference procedure will be presented in Section 5.2. First, for corpora which contain irrelevant snippets, we may introduce an additional word distribution $\theta_I$ and word topic `Ignore` to allow the model to ignore certain snippets or pieces of snippets altogether. Second, if it is possible to acquire part of speech tags for the snippets, using these as an extra piece of information is quite beneficial. Finally, for corpora where every entity is expected to share the same aspects, the model can be altered to use the same set of aspect distributions for all entities.

### 4.2.1 IGNORING SNIPPETS

When snippet data is automatically extracted, it may be noisy, and some snippets may violate our initial assumptions of having one aspect and one value. For example, we find some snippets which were mistakenly extracted that have neither aspect nor value. These extraneous snippets may be difficult to identify a priori. To compensate for this, we modify the model to allow partial or entire snippets to be ignored through the addition of a global unigram distribution, namely the Ignore distribution $\theta_I$. This distribution is drawn from a symmetric Dirichlet with concentration parameter $\lambda_I$.

The Ignore distribution differs from the Background distribution in that it includes both common and uncommon words. It is intended to select whole snippets or large portions of snippets, so some words may overlap with the Background distribution and other distributions. In order to successfully incorporate this distribution into our model, we must allow the word topic indicator $Z_W^{i,j,w}$ to consider the Ignore topic $I$. Additionally, to ensure that it selects long segments of text, we give a large boost to the prior of the `Ignore Ignore` sequence in the transition distribution $\Lambda$, similar to the boost for self-transitions.

### 4.2.2 PART-OF-SPEECH TAGS

Part-of-speech tags can provide valuable evidence in determining which snippet words are drawn from each distribution. For example, aspect words are often nouns, as they represent concrete properties or concepts in a domain. Likewise, in some domains, value words describe aspects and therefore tend to be expressed as numbers or adjectives.

This intuition can be directly incorporated into the model in the form of additional outputs. Specifically, we modify our HMM to produce both words and tags. Additionally, we define distributions over tags $\eta_A^a$, $\eta_V^v$, and $\eta_B$, similar to the corresponding unigram distributions.

### 4.2.3 SHARED ASPECTS

When domains are very regular, and every entity is expected to express aspects from a consistent set, it is beneficial to share aspect information across entities. For example, in a medical domain, the same general set of lab tests and physical exam categories are run on all patients. Note that this is quite unlike the restaurant review case, where each restaurant's aspects are completely different (e.g., *pizza*, *curry*, *scones*, and so on).

Sharing aspects in this way can be accomplished by modifying the aspect distributions $\theta_A^{i,a}$ to become global distributions $\theta_A^a$. Likewise, aspect-value multinomials $\phi^{i,a}$ become shared across all entities as as $\phi^a$. Treatment of the aspect multinomials depend on the domain properties. If the distribution over aspects is expected to be the same across all entities, it can also be made global; however, if each individual entity is expected to exhibit variation in the number of snippets related to each aspect, they should be kept as entity-specific. For example, reviews for a set of cell phones may be expected to focus on varying parts, depending on what is most unique or problematic about those phones. A graphical description of these changes compared to the original model is shown in Figure 6.

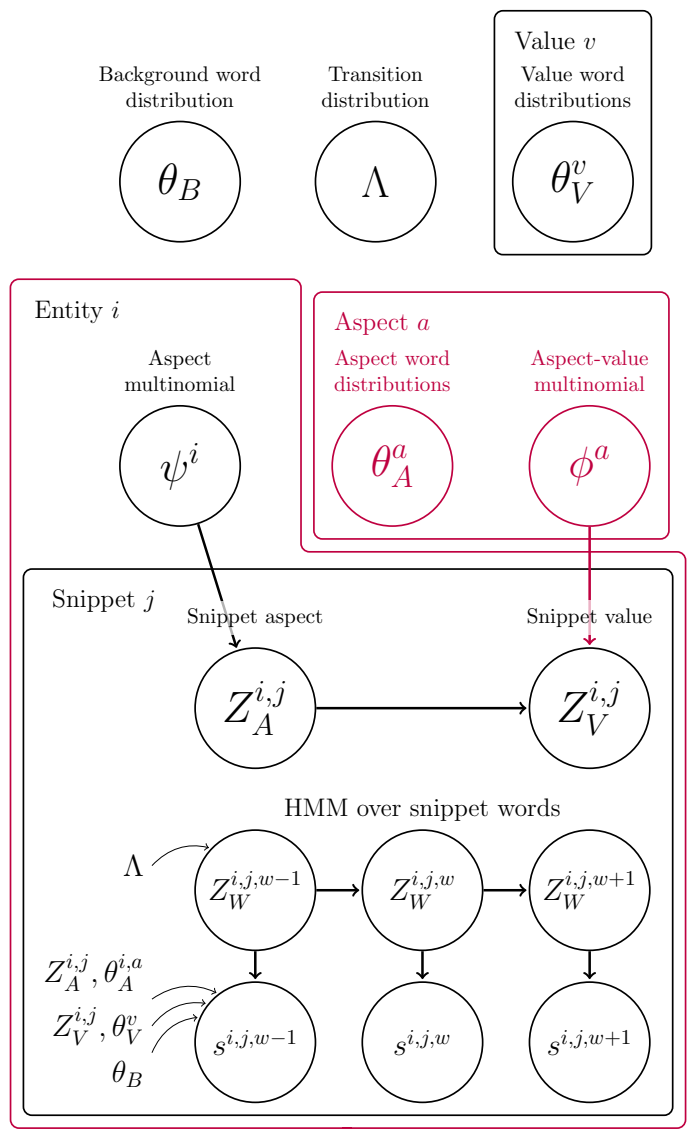

Figure 6: A graphical description of the model with shared aspects presented in Section 4.2. Note the similarities to Figure 5; however in this version, aspects are shared for the entire corpus, rather than being entity-specific. It would also be possible to share the aspect multinomial corpus-wide; in that case it would indicate that all entities share the same general distribution over aspects, while in this version the individual entities are allowed to have completely different distributions.

**Mean-field Factorization**

$$Q\left(\theta_B, \boldsymbol{\theta_V}, \Lambda, \boldsymbol{\theta_A}, \psi, \phi, \boldsymbol{Z}\right)$$

$$= q\left(\theta_B\right) q\left(\Lambda\right) \left(\prod_{v=1}^{N} q\left(\theta_V^v\right)\right) \left(\prod_{i} q\left(\psi^i\right) \left(\prod_{a=1}^{K} q\left(\theta_A^{i,a}\right) q\left(\phi^{i,a}\right)\right) \left(\prod_{j} q\left(Z_V^{i,j}\right) q\left(Z_A^{i,j}\right) \prod_{w} q\left(Z_W^{i,j,w}\right)\right)\right)$$

**Snippet Aspect Indicator**

$$\log q(Z_A^{i,j} = a) \propto \mathbb{E}_{q(\psi^i)} \log \psi^i(a) + \sum_w q(Z_W^{i,j,w} = A)\mathbb{E}_{q(\theta_A^{i,a})} \log \theta_A^{i,a}(s^{i,j,w}) + \sum_{v=1}^{N} q(Z_V^{i,j} = v)\mathbb{E}_{q(\phi^{i,a})} \log \phi^{i,a}(v)$$

**Snippet Value Type Indicator**

$$\log q(Z_V^{i,j} = v) \propto \sum_a q(Z_A^{i,j} = a)\mathbb{E}_{q(\phi^{i,a})} \log \phi^{i,a}(v) + \sum_w q(Z_W^{i,j,w} = V)\mathbb{E}_{q(\theta_V^v)} \log \theta_V^v(s^{i,j,w})$$

**Word Topic Indicator**

$$\log q(Z_W^{i,j,w} = A) \propto \log P(Z_W = A) + \mathbb{E}_{q(\Lambda)} \log\left(\Lambda\left(Z_W^{i,j,w-1}, A\right)\Lambda\left(A, Z_W^{i,j,w+1}\right)\right) + \sum_a q(Z_A^{i,j} = a)\mathbb{E}_{q(\theta_A^{i,a})} \log \theta_A^{i,j}(s^{i,j,w})$$

$$\log q(Z_W^{i,j,w} = V) \propto \log P(Z_W = V) + \mathbb{E}_{q(\Lambda)} \log\left(\Lambda\left(Z_W^{i,j,w-1}, V\right)\Lambda\left(V, Z_W^{i,j,w+1}\right)\right) + \sum_v q(Z_V^{i,j} = v)\mathbb{E}_{q(\theta_V^v)} \log \theta_V^v(s^{i,j,w})$$

$$\log q(Z_W^{i,j,w} = B) \propto \log P(Z_W = B) + \mathbb{E}_{q(\Lambda)} \log\left(\Lambda\left(Z_W^{i,j,w-1}, B\right)\Lambda\left(B, Z_W^{i,j,w+1}\right)\right) + \mathbb{E}_{q(\theta_B)} \log \theta_B(s^{i,j,w})$$

Figure 7: The mean-field variational algorithm used during learning and inference to obtain posterior predictions over snippet properties and attributes, as described in Section 5. Mean-field inference consists of updating each of the latent variable factors as well as a straightforward update of latent parameters in round robin fashion.

## 5. Inference

The goal of inference in this model is to predict the aspect and value for each snippet $i$ and product $j$, given the text of all observed snippets, while marginalizing out the remaining hidden parameters:

$$P(Z_A^{i,j}, Z_V^{i,j} | \mathbf{s})$$

We accomplish this task using variational inference (Blei, Ng, & Jordan, 2003). Specifically, the goal of variational inference is to find a tractable approximation $Q(\cdot)$ to the full posterior of the model.

$$P(\theta_B, \boldsymbol{\theta_V}, \Lambda, \boldsymbol{\theta_A}, \psi, \phi, \boldsymbol{Z} | \mathbf{s}) \approx Q(\theta_B, \boldsymbol{\theta_V}, \Lambda, \boldsymbol{\theta_A}, \psi, \phi, \boldsymbol{Z})$$

For our model, we assume a full mean-field factorization of the variational distribution, shown in Figure 7. This variational approximation is defined as a product of factors $q(\cdot)$, which are assumed to be independent. This approximation allows for tractable inference of each factor individually. To obtain the closest possible approximation, we attempt to set the $q(\cdot)$ factors to minimize the KL divergence to the true model posterior:

$$\underset{Q(\cdot)}{\arg\min} \, KL(Q(\theta_B, \boldsymbol{\theta_V}, \Lambda, \boldsymbol{\theta_A}, \boldsymbol{\psi}, \boldsymbol{\phi}, \boldsymbol{Z}) \| P(\theta_B, \boldsymbol{\theta_V}, \Lambda, \boldsymbol{\theta_A}, \boldsymbol{\psi}, \boldsymbol{\phi}, \boldsymbol{Z} | \mathbf{s}))$$

## 5.1 Optimization

We optimize this objective using coordinate descent on the $q(\cdot)$ factors. Concretely, we update each factor by optimizing the above criterion with all other factors fixed to their current values:

$$q(\cdot) \leftarrow \mathbb{E}_{Q/q(\cdot)} \log P(\theta_B, \boldsymbol{\theta_V}, \Lambda, \boldsymbol{\theta_A}, \boldsymbol{\psi}, \boldsymbol{\phi}, \boldsymbol{Z}, \mathbf{s})$$

A summary of the variational update equations is given in Figure 7, and a graphical representation of the involved variables for each step is presented in Figure 8. Here, we will present the update for each factor.

### 5.1.1 SNIPPET ASPECT INDICATOR

First, we consider the update for the snippet aspect indicator, $Z_A^{i,j}$ (Figure 8a):

$$\log q(Z_A^{i,j} = a) \propto \mathbb{E}_{q(\psi^i)} \log \psi^i(a) \tag{1a}$$

$$+ \sum_w q(Z_W^{i,j,w} = A) \mathbb{E}_{q(\theta_A^{i,a})} \log \theta_A^{i,a}(s^{i,j,w}) \tag{1b}$$

$$+ \sum_{v=1}^N q(Z_V^{i,j} = v) \mathbb{E}_{q(\phi^{i,a})} \log \phi^{i,a}(v) \tag{1c}$$

The optimal aspect for a particular snippet depends on three factors. First, we include the likelihood of discussing each aspect $a$ (Eqn. 1a). As mentioned earlier, this encodes the prior probability that some aspects are discussed more frequently than others. Second, we examine the likelihood of a particular aspect based on the words in the snippet (Eqn. 1b). For each word which is identified as an aspect word, we add the probability that it discusses this aspect. Third, we determine the compatibility of the chosen aspect type with the current aspect (Eqn. 1c). For example, if we know the value type is most likely an integer, the assigned aspect should accept integers.

### 5.1.2 SNIPPET VALUE TYPE INDICATOR

Next, we consider the update for the snippet value type indicator, $Z_V^{i,j}$ (Figure 8b):

$$\log q(Z_V^{i,j} = v) \propto \sum_a q(Z_A^{i,j} = a) \mathbb{E}_{q(\phi^{i,a})} \log \phi^{i,a}(v) \tag{2a}$$

$$+ \sum_w q(Z_W^{i,j,w} = V) \mathbb{E}_{q(\theta_V^v)} \log \theta_V^v(s^{i,j,w}) \tag{2b}$$

The best value type for a snippet depends on two factors. First, like the snippet aspect indicator, we must take into consideration the compatibility between snippet aspect and value type (Eqn. 2a). Second, for each word identified as a value word, we include the likelihood that it comes from the given value type.

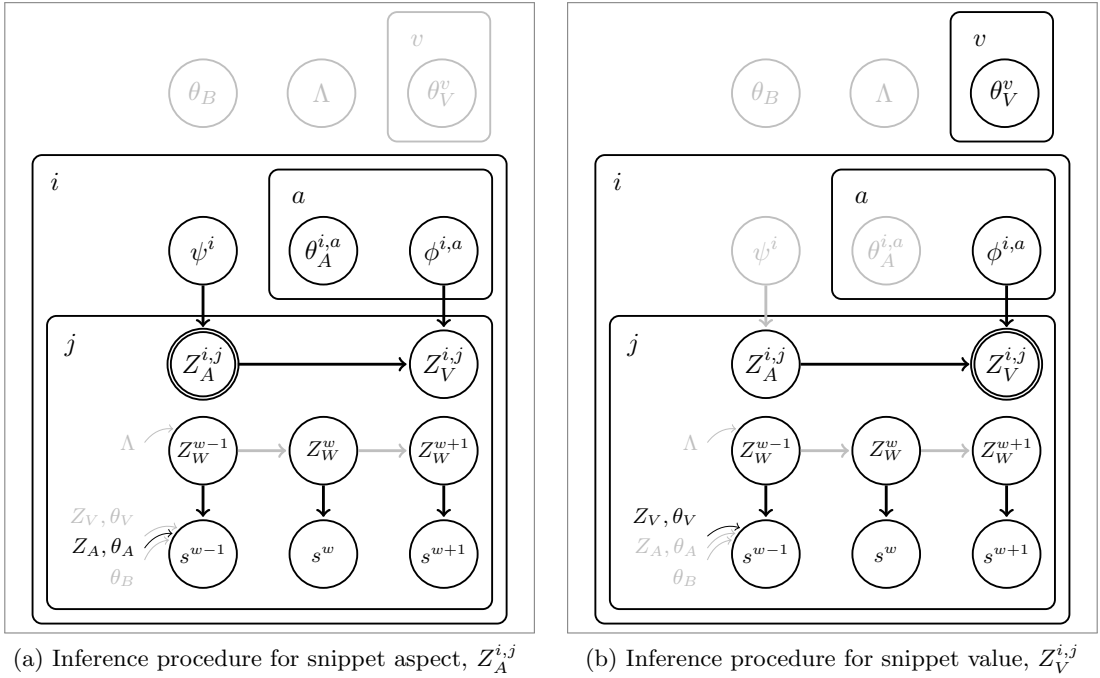

(a) Inference procedure for snippet aspect, $Z_A^{i,j}$

(b) Inference procedure for snippet value, $Z_V^{i,j}$

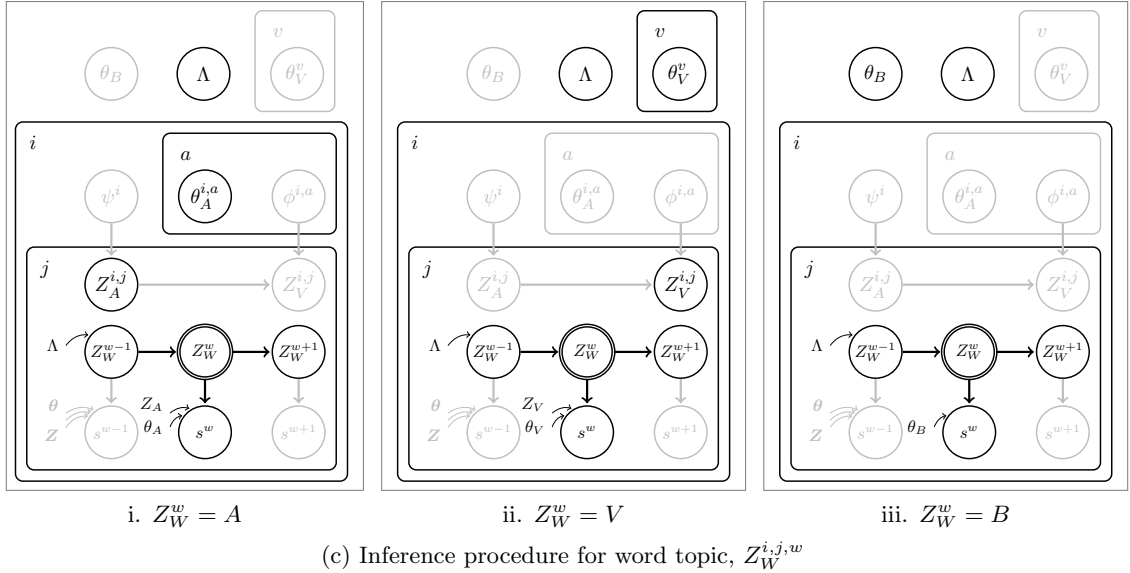

i. $Z_W^w = A$

ii. $Z_W^w = V$

iii. $Z_W^w = B$

(c) Inference procedure for word topic, $Z_W^{i,j,w}$

Figure 8: Variational inference update steps for each latent variable. The latent variable currently being updated is shown in a double circle, and the other variables relevant to the update are highlighted in black. Those variables which have no impact on the update are grayed out. Note that for snippet aspect (a) and snippet value type (b), the update takes the same form for each possible aspect or value type. However, for word topic (c), the update is not symmetric as the relevant variables are different for each possible word topic.

### 5.1.3 Word Topic Indicator

Finally, we consider the update for the word topic indicators, $Z_W^{i,j,w}$ (Figure 8c). Unlike the previous indicators, each possible topic has a slightly different equation, as we must marginalize over all possible aspects and value types.

$$\log q\big(Z_W^{i,j,w} = A\big) \propto \log P\big(Z_W = A\big) + \mathbb{E}_{q(\Lambda)} \log\Big(\Lambda\big(Z_W^{i,j,w-1}, A\big)\Lambda\big(A, Z_W^{i,j,w+1}\big)\Big)$$
$$+ \sum_a q\big(Z_A^{i,j} = a\big)\mathbb{E}_{q(\theta_A^{i,a})} \log\theta_A^{i,j}\big(s^{i,j,w}\big) \tag{3a}$$

$$\log q\big(Z_W^{i,j,w} = V\big) \propto \log P\big(Z_W = V\big) + \mathbb{E}_{q(\Lambda)} \log\Big(\Lambda\big(Z_W^{i,j,w-1}, V\big)\Lambda\big(V, Z_W^{i,j,w+1}\big)\Big)$$
$$+ \sum_v q\big(Z_V^{i,j} = v\big)\mathbb{E}_{q(\theta_V^v)} \log\theta_V^v\big(s^{i,j,w}\big) \tag{3b}$$

$$\log q\big(Z_W^{i,j,w} = B\big) \propto \log P\big(Z_W = B\big) + \mathbb{E}_{q(\Lambda)} \log\Big(\Lambda\big(Z_W^{i,j,w-1}, B\big)\Lambda\big(B, Z_W^{i,j,w+1}\big)\Big)$$
$$+ \mathbb{E}_{q(\theta_B)} \log\theta_B\big(s^{i,j,w}\big) \tag{3c}$$

The update for each topic is composed of the prior probability of having that topic, transition probabilities using this topic, and the probability of the word coming from the appropriate unigram distribution, marginalized over all possibilities for snippet aspect and value indicators.

### 5.1.4 Parameter Factors

Updates for the parameter factors under variational inference are derived through simple counts of the latent variables $Z_A$, $Z_V$, and $Z_W$. Note that these do include partial counts; if a particular snippet has aspect probability $P(Z_A^{i,j} = a_1) = 0.35$, it would contribute 0.35 count to $\psi^i(a_1)$.

### 5.1.5 Algorithm Details

Given this set of update equations, the update procedure is straightforward. First, iterate over the corpus computing the updated values for each random variable, then do a batch update for all factors simultaneously. This update algorithm is run to convergence. In practice, convergence is achieved by the 50th iteration, so the algorithm is quite efficient.

Note that the batch update means each update is computed using the values from the previous iteration, unlike Gibbs sampling which uses updated values as it runs through the corpus. This difference allows the variational update algorithm to be parallelized, yielding a nice efficiency boost. Specifically, to parallelize the algorithm, we simply split the set of entities evenly among processors. Updates for entity-specific factors and variables are computed during the pass through the data, and updates for global factors are collected and combined at the end of each pass.

## 5.2 Inference for Model Extensions

As discussed in Section 4.2, we can add additional components to the model to improve performance for data with certain attributes. Here, we briefly discuss the modifications to the inference equations for each extension.

### 5.2.1 Ignoring Snippets

The main modifications to the model for this extension are the addition of the unigram distribution $\theta_I$ and word topic $I$, which can be chosen by $Z_W$. The update equation for $Z_W$ is modified by the addition of the following:

$$\log q(Z_W^{i,j,w} = I) \propto \log P(Z_W = I) + \mathbb{E}_{q(\theta_I)} \log \theta_I(s^{i,j,w})$$

As in the other pieces of this equation (Eqn. 3), this is composed of the prior probability for the word topic $I$ and the likelihood that this word is generated by $\theta_I$.

In addition, the transition distribution $\Lambda$ must be updated to include transition probabilities for $I*$ and $*I$. As mentioned earlier, the $II$ transition receives high weight, while all other transitions to and from $I$ receive very low weight.

### 5.2.2 Part-of-Speech Tags

To add part of speech tags, the model is updated to include part-of-speech distributions $\eta_A$, $\eta_V$, and $\eta_B$, one for each word topic. Note that unlike the unigram distributions $\theta_A^{i,a}$ and $\theta_V^v$, the corresponding tag distributions are not dependent on snippet entity, aspect, or value. These are included and referenced in the updates for $Z_W$ as follows:

$$\log q(Z_W^{i,j,w} = A) \propto \log P(Z_W = A) + \mathbb{E}_{q(\Lambda)} \log\Big(\Lambda(Z_W^{i,j,w-1}, A)\Lambda(A, Z_W^{i,j,w+1})\Big)$$
$$+ \mathbb{E}_{q(\eta_A)} \log \eta_A(t^{i,j,w}) + \sum_a q(Z_A^{i,j} = a)\mathbb{E}_{q(\theta_A^{i,a})} \log \theta_A^{i,j}(s^{i,j,w})$$

$$\log q(Z_W^{i,j,w} = V) \propto \log P(Z_W = V) + \mathbb{E}_{q(\Lambda)} \log\Big(\Lambda(Z_W^{i,j,w-1}, V)\Lambda(V, Z_W^{i,j,w+1})\Big)$$
$$+ \mathbb{E}_{q(\eta_V)} \log \eta_V(t^{i,j,w}) + \sum_v q(Z_V^{i,j} = v)\mathbb{E}_{q(\theta_V^v)} \log \theta_V^v(s^{i,j,w})$$

$$\log q(Z_W^{i,j,w} = B) \propto \log P(Z_W = B) + \mathbb{E}_{q(\Lambda)} \log\Big(\Lambda(Z_W^{i,j,w-1}, B)\Lambda(B, Z_W^{i,j,w+1})\Big)$$
$$+ \mathbb{E}_{q(\eta_B)} \log \eta_B(t^{i,j,w}) + \mathbb{E}_{q(\theta_B)} \log \theta_B(s^{i,j,w})$$

Here, we define $\mathbf{t}$ as the set of all tags and $t^{i,j,w}$ as the tag corresponding to the word $s^{i,j,w}$.

### 5.2.3 Shared Aspects

A global set of shared aspects is a simplification of the model in that it reduces the total number of parameters. This model redefines aspect distributions to be $\theta_A^a$ and aspect-value multinomials to be $\phi^a$. Depending on domain, it may also redefine the aspect multinomial to be $\psi$. The resulting latent variable update equations are the same; only the parameter factor updates are changed. Rather than collecting counts over snippets describing a single entity, counts are collected across the corpus.

## 6. Experiments

We perform experiments on two tasks. First, we test our full model on joint prediction of both aspect and sentiment on a corpus of review data. Second, we use a simplified version of the model designed to identify aspects only on a corpus of medical summary data. These domains are structured quite differently, and therefore present very different challenges to our model.

### 6.1 Joint Identification of Aspect and Sentiment

Our first task is to test our full model by jointly predicting both aspect and sentiment on a collection of restaurant review data. Specifically, we would like to dynamically select a set of relevant aspects for each restaurant, identify the snippets which correspond to each aspect, and recover the polarity of each snippet individually and each aspect as a whole. We perform three experiments to evaluate our model's effectiveness. First, we test the quality of learned aspects by evaluating the predicted snippet clusters. Second, we assess the quality of the polarity classification. Third, we examine per-word labeling accuracy.

#### 6.1.1 Data Set

Our data set for this task consists of snippets selected from Yelp restaurant reviews by our previous system (Sauper et al., 2010). The system is trained to extract snippets containing short descriptions of user sentiment towards some aspect of a restaurant.[4] For the purpose of this experiment, we select only the snippets labeled by that system as referencing *food*. In order to ensure that there is enough data for meaningful analysis, we ignore restaurants that have fewer than 20 snippets across all reviews. While our model can easily operate on restaurants with fewer snippets, we want to ensure that the cases we select for evaluation are nontrivial; i.e., that there are a sufficient number of snippets in each cluster to make a valid comparison. There are 13,879 snippets in total, taken from 328 restaurants in and around the Boston/Cambridge area. The average snippet length is 7.8 words, and there are an average of 42.1 snippets per restaurant. We use the MXPOST tagger (Ratnaparkhi, 1996) to gather POS tags for the data. Figure 9 shows some example snippets.

For this domain, the value distributions consist of one positive and one negative distribution. These are seeded using 42 and 33 seed words respectively. Seed words are hand-selected based on the restaurant review domain; therefore, they include domain-specific words such as *delicious* and *gross*. A complete list of seed words is included in Table 2.

#### 6.1.2 Domain Challenges and Modeling Techniques

This domain presents two challenging characteristics for our model. First, there are a wide variety of restaurants within our domain, including everything from high-end Asian fusion cuisine to greasy burger fast food places. If we were to try to represent these using a single shared set of aspects, the number of aspects required would be immense, and it would be extremely difficult for our model to make fine-grained distinctions between them. By defining aspects separately for each restaurant as mentioned in Section 4, we can achieve the proper granularity of aspects for each individual restaurant without an overwhelming or overlapping selection of choices. For example, the model is able to distinguish that an Italian restaurant may need only a single *dessert* aspect, while a bakery requires separate *pie*, *cake*, and *cookie* aspects.

---

4. For exact training procedures, please reference that paper.

| Positive | | | | Negative | | |
|---|---|---|---|---|---|---|
| amazing | awesome | best | delicious | average | awful | bad |
| delightful | divine | enjoy | excellent | bland | boring | confused |
| extraordinary | fantastic | fav | favorite | disappointed | disgusting | dry |
| flavorful | free | fresh | fun | expensive | fatty | greasy |
| generous | good | great | happy | gross | horrible | inedible |
| heaven | huge | incredible | interesting | lame | less | mediocre |
| inexpensive | love | nice | outstanding | meh | mushy | overcooked |
| perfect | phenomenal | pleasant | quality | poor | pricey | salty |
| recommend | rich | sleek | stellar | tacky | tasteless | terrible |
| stimulating | strong | tasty | tender | tiny | unappetizing | underwhelming |
| wonderful | yummy | | | uninspiring | worse | worst |

Table 2: Seed words used by the model for the restaurant corpus, 42 positive words and 33 negative words in total. These words are manually selected for this data set.

Second, while there are usually a fairly cohesive set of words which refer to any particular aspect (e.g., the *pizza* aspect might be commonly be seen with the words *slice*, *pepperoni*, and *cheese*), there are a near-unlimited set of potential sentiment words. This is especially pronounced in the social media domain where there are many novel words used to express sentiment (e.g., *deeeeeeelish* as a substitute for *delicious*). As mentioned in Section 4, the part-of-speech and transition components of the model helps to identify which unknown words are likely to be sentiment words; however, we additionally need to identify the polarity of their sentiment. To do this, we can leverage the aspect-value multinomial, which represents the likelihood of positive or negative sentiment for a particular aspect. If most of the snippets about a given aspect are positive, it is likely that the word *deeeeeeelish* represents positive sentiment as well.

### 6.1.3 CLUSTER PREDICTION

The goal of this task is to evaluate the quality of aspect clusters; specifically the $Z_A^{i,j}$ variable in Section 4. In an ideal clustering, the predicted clusters will be cohesive (i.e., all snippets predicted to discuss a given aspect are related to each other) and comprehensive (i.e., all snippets which discuss an aspect are selected as such). For example, a snippet will be assigned the aspect *pizza* if and only if that snippet mentions some aspect of pizza, such as its crust, cheese, or toppings.

**Annotation** For this experiment, we use a set of gold clusters on the complete sets of snippets from 20 restaurants, 1026 snippets in total (an average of 51.3 snippets per restaurant). Cluster annotations were provided by graduate students fluent in English. Each annotator was provided with a complete set of snippets for a particular restaurant, then asked to cluster them naturally. There were 199 clusters in total, which yields an average

> The **noodles** and the **meat** were actually [+]**pretty good**.
> I [+]**recommend** the **chicken noodle pho**.
> The **noodles** were [-]**soggy**.
> The **chicken pho** was also [+]**good**.

> The **spring rolls** and **coffee** were [+]**good**, though.
> The **spring roll wrappers** were a [-]**little dry tasting**.
> My [+]**favorites** were the **crispy spring rolls**.
> The **Crispy Tuna Spring Rolls** are [+]**fantastic**!

> The **lobster roll** my mother ordered was [-]**dry** and [-]**scant**.
> The **portabella mushroom** is my [+]**go-to** **sandwich**.
> The **bread** on the **sandwich** was [-]**stale**.
> The slice of **tomato** was [-]**rather measly**.

> The **shumai** and **california maki sushi** were [+]**decent**.
> The **spicy tuna roll** and **eel roll** were [+]**perfect**.
> The **rolls** with **spicy mayo** were [-]**not so great**.
> I [+]**love** **Thai rolls**.

Figure 9: Example snippets from our data set, grouped according to aspect. Aspect words are underlined and colored blue, NEGATIVE value words are labeled **-** and colored red, and POSITIVE value words are labeled **+** and colored green. The grouping and labeling are *not* given in the data set and must be learned by the model.

of 10.0 clusters per restaurant. These annotations are high-quality; the average annotator agreement is 81.9 by the MUC evaluation metric (described in detail below). While we could define a different number of clusters for each restaurant by varying the number of aspect distributions, for simplicity we ask both baseline systems and our full model to produce 10 aspect clusters per restaurant, matching the average annotated number. Varying the number of clusters will simply cause existing clusters to merge or split; there are no large or surprising changes in clustering.

**Baseline** We use two baselines for this task, both using a clustering algorithm weighted by TF*IDF as implemented by the publicly available CLUTO package (Karypis, 2002),[5] using agglomerative clustering with the cosine similarity distance metric (Chen, Branavan, Barzilay, & Karger, 2009; Chen, Benson, Naseem, & Barzilay, 2011).

The first baseline, CLUSTER-ALL, clusters over entire snippets in the data set. This baseline will put a strong connection between things which are lexically similar. Because our model only uses aspect words to tie together clusters, this baseline may capture correlations between words which our model does not correctly identify as aspect words.

---

5. Available at `http://glaros.dtc.umn.edu/gkhome/cluto/cluto/overview`.

|  | Precision | Recall | F1 |
|---|---|---|---|
| Cluster-All | 57.3 | 60.1 | 58.7 |
| Cluster-Noun | 68.6 | 70.5 | 69.5 |
| Our model | **74.3** | **85.3** | **79.4** |

Table 3: Results using the MUC metric on cluster prediction for the joint aspect and value identification task. While MUC has a deficiency in that putting everything into a single cluster will artificially inflate the score, all models are set to use the same number of clusters. Note that for this task, the Cluster-Noun significantly outperforms the Cluster-All baseline, indicating that part of speech is a crucial piece of information for this task.

The second baseline, Cluster-Noun, works over only the nouns from the snippets. Each snippet is POS-tagged using MXPOST (Ratnaparkhi, 1996),[6] and any non-noun (i.e., not NN, NNS, NNP, or NNPS) words are removed. Because we expect that most aspects contain at least one noun, this acts as a proxy for the aspect identification in our model.

**Metric** We use the MUC cluster evaluation metric for this task (Vilain, Burger, Aberdeen, Connolly, & Hirschman, 1995). This metric measures the number of cluster merges and splits required to recreate the gold clusters given the model's output. Therefore, it can concisely show how accurate our clusters are as a whole. While it would be possible to artificially inflate the score by putting everything into a single cluster, the parameters on our model and the likelihood objective are such that the model prefers to use all available clusters, the same number as the baseline system.

**Results** Results for our cluster prediction task are in Table 3. Our model shows strong performance over each baseline, for a total error reduction of 32% over the Cluster-Noun baseline and 50% over the Cluster-All baseline. The most common cause of poor cluster choices in the baseline systems is their inability to distinguish which words are relevant aspect words. For example, in the Cluster-All baseline, if many snippets use the word *delicious*, there may end up being a cluster based on that alone. The Cluster-Noun baseline is able to avoid some of these pitfalls thanks to its built-in filter. It is able to avoid common value words such as adjectives and also focus on what seems to be the most concrete portion of the aspect (e.g., *blackened* **chicken**); however, it still cannot make the correct distinctions where these assumptions are broken. Because our model is capable of distinguishing which words are aspect words (i.e., words relevant to clustering), it can choose clusters which make more sense overall.

### 6.1.4 Sentiment Analysis

We evaluate the system's predictions of snippet sentiment using the predicted posterior over the value distributions for the snippet (i.e., $Z_A^{i,j}$). For this task, we consider the binary judgment to be simply the one with higher value in $q(Z_A^{i,j})$ (see Section 5). The goal of this task is to evaluate whether our model correctly distinguishes the sentiment of value words.

---

6. Available at `http://www.inf.ed.ac.uk/resources/nlp/local_doc/MXPOST.html`.

|  | Accuracy |
|---|---|
| Majority | 60.7 |
| Discriminative-Small | 74.1 |
| Seed | 78.2 |
| Discriminative-Large | 80.4 |
| Our model | **82.5** |

Table 4: Sentiment prediction accuracy of our model compared to the Discriminative and Seed baselines, as well as Majority representing the majority class (Positive) baseline. One advantage of our system is its ability to distinguish aspect words from sentiment words in order to restrict judgment to only the relevant terms; another is the leverage that it gains from biasing unknown sentiment words to follow the polarity observed in other snippets relating to the same aspect.

**Annotation**  For this task, we use a set of 662 randomly selected snippets from the Yelp reviews which express opinions. To get a clear result, this set specifically excludes neutral, mixed, or potentially ambiguous snippets such as *the fries were too salty but tasty* or *the blackened chicken was very spicy*, which make up about 10% of the overall data. This set is split into a training set of 550 snippets and a test set of 112 snippets, then each snippet is manually labeled positive or negative. For one baseline, we use the set of positive and negative seed words which were manually chosen for our model, shown in Table 2. Note that as before, our model has access to the full corpus of unlabeled data plus the seed words, but no labeled examples.

**Baseline**  We use two baselines for this task, one based on a standard discriminative classifier and one based on the seed words from our model.

The Discriminative baseline for this task is a standard maximum entropy discriminative binary classifier[7] over unigrams. Given enough snippets from enough unrelated aspects, the classifier should be able to identify that words like *great* indicate positive sentiment and those like *bad* indicate negative sentiment, while words like *chicken* are neutral and have no effect. To illustrate the effect of training size, we include results for Discriminative-Small, which uses 100 training examples, and Discriminative-Large, which uses 550 training examples.

The Seed baseline simply counts the number of words from the same positive and negative seed lists used by the model, $V_{seed_+}$ and $V_{seed_-}$, as listed in Table 2. If there are more words from $V_{seed_+}$, the snippet is labeled positive, and if there are more words from $V_{seed_-}$, the snippet is labeled negative. If there is a tie or there are no seed words, we split the prediction. Because the seed word lists are manually selected specifically for restaurant reviews (i.e., they contain food-related sentiment words such as *delicious*), this baseline should perform well.

**Results**  The overall sentiment classification accuracy of each system are shown in Table 4). Our model outperforms both baselines. The obvious flaw in the Seed baseline is the inability to pre-specify every possible sentiment word. It does perform highly, due to its tailoring for the restaurant domain and good coverage of the most frequent words (e.g., *delicious*, *good*, *great*), but the performance of our model indicates that it can generalize beyond these seed words.

---

7. Available at `https://github.com/lzhang10/maxent`.

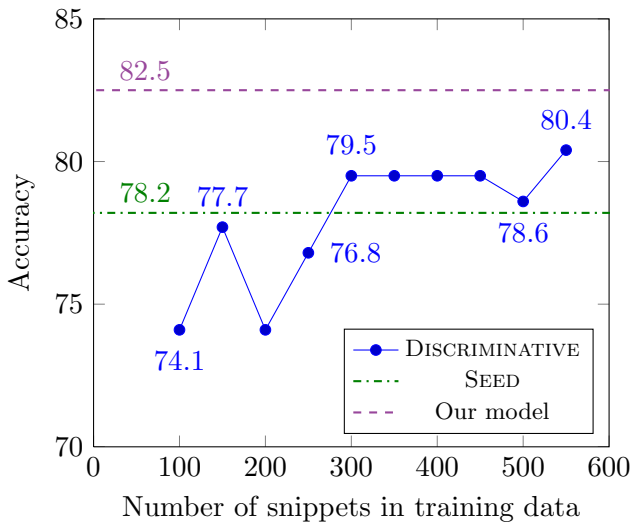

Figure 10: DISCRIMINATIVE baseline performance as the number of training examples increases. While performance generally increases, there are some inconsistencies. The main issue with this baseline is that it needs to see examples of words in training data before it can improve; this phenomenon can be seen at the plateau in this graph.

The DISCRIMINATIVE-LARGE outperforms the SEED baseline on this test set; however, given the smaller training set of DISCRIMINATIVE-SMALL, it performs worse. The training curve of the DISCRIMINATIVE baseline is shown in Figure 10. While the DISCRIMINATIVE baseline system can correctly identify the polarity of statements containing information it has seen in the past, it has two main weaknesses. First, every sentiment word must have been present in training data. For example, in our test data, *rancid* appears in a negative sentence; however, it does not appear in the training data, so the model labels the example incorrectly. This is problematic, as there is no way to find training data for every possible sentiment word, especially in social media data where novel words and typos are a frequent occurrence. Our model's ability to generalize about the polarity of snippets describing a particular aspect allows it to predict sentiment values for words of unknown polarity. For example, if there are already several positive snippets describing a particular aspect, the system can guess that a snippet with unknown polarity will likely also be positive.

### 6.1.5 PER-WORD LABELING ACCURACY

The goal of this task is to evaluate whether each word is correctly identified as an aspect word, value word, or background word. This distinction is crucial in order to achieve correctness of both clustering and sentiment analysis, so errors here may help us identify weaknesses of our model.

> The **rolls** also were **n't very well made** .
> The **pita** was **beyond dry** and **tasted like cardboard** !
> The Falafel King has the **best falafel** !
> The **rolls with spicy mayo** were **not so good** .
> Ordered the **spicy tuna** and **california roll** – they were **amazing** !

Table 5: Correct annotation of a set of phrases containing elements which may be confusing, on which annotators are tested before they are allowed to annotate the actual test data. Aspect words are colored blue and underlined; value words are colored orange and underlined with a wavy line. Some common mistakes include: annotating *n't* as background (because it is attached to the background word *was*), annotating *cardboard* as an aspect because it is a noun, annotating *Falafel King* as an aspect because it is in subject position.

**Annotation**  Per-word annotation is acquired from Mechanical Turk. The per-word labeling task seems difficult for some Turk annotators, so we implement a filtering procedure to ensure that only high-quality annotators are allowed to submit results. Specifically, we ask annotators to produce labels for a set of "difficult" phrases with known labels (shown in Table 5). Those annotators who successfully produced correct or mostly-correct annotations are allowed to access the annotation tasks containing new phrases. Each of these unknown tasks is presented to 3 annotators, and the majority label is taken for each word. In total, we test on 150 labeled phrases, for a total of 7,401 labeled words.

**Baseline**  The baseline for this task relies again on the intuition that part-of-speech is a useful proxy for aspect and value identification. We know that aspects usually represent concrete entities, so they are often nouns, and value words are descriptive or counting, so they are often adjectives or adverbs. Therefore, we again use the MXPOST tagger to find POS for each word in the snippet. For the main baseline, Tags-Full, we assign each noun (`NN*`) an aspect label, and each numeral, adjective, adverb, or verb participle (`CD`, `RB*`, `JJ*`, `VBG`, `VBN`) a value label. For comparison, we also present results for a smaller tagset, Small-Tags, labeling only nouns (`NN*`) as aspect and adjectives (`JJ*`) as values. Note that each of the tags added in the Tags-Full baseline are beneficial to the baseline's score.

**Tree expansion**  Because our full model and the baselines are all designed to pick out relevant individual words rather than phrases, they may not correspond well to the phrases which humans have selected as relevant. Therefore, we also evaluate on a set of expanded labels identified with parse trees from the Stanford Parser (Klein & Manning, 2003).[8] Specifically, for each non-background word, we identify the largest containing noun phrase (for both aspects and values) or adjective or adverb phrase (for values only) which does not also contain oppositely-labeled words. For example, in the noun phrase *blackened chicken*, if *chicken* was labeled as an aspect word and *blackened* was labeled as a background word, both will now be labeled as aspect words. However, in the noun phrase *tasty chicken* where "tasty" is already labeled as a value, the label will not be changed and no further expansion will be attempted. As a final heuristic step, any punctuation, determiners, and conjunctions which would be newly labeled as aspect or value words are ignored and kept as background words. The steps of this procedure with an illustrative example are shown in Figure 11.

---

8. Available at `http://nlp.stanford.edu/software/lex-parser.shtml`.

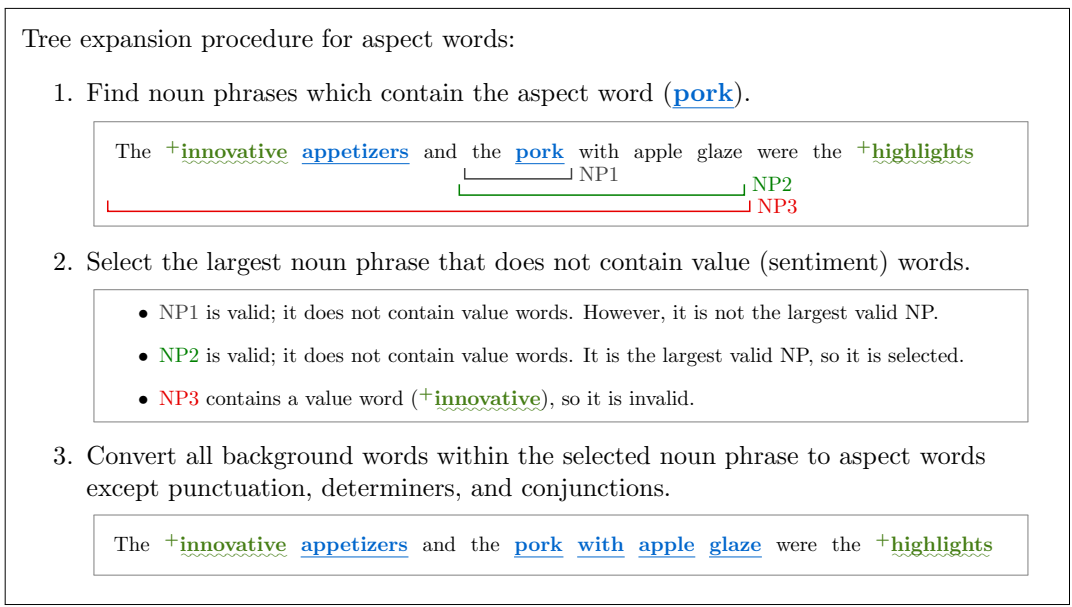

Figure 11: The tree expansion procedure for value words, with an example snippet. The procedure is similar for aspect words, except adjective phrases and adverb phrases are also considered for expansion.

|  | Aspect | | | Value | | |
|---|---|---|---|---|---|---|
|  | Precision | Recall | F$_1$ | Precision | Recall | F$_1$ |
| Tags-Small | 79.9 | 79.5 | **79.7** | 78.5 | 45.0 | 57.2 |
| Tree | 74.0 | **83.0** | 78.2 | **79.2** | 57.4 | 66.5 |
| Tags-Full | 79.9 | 79.5 | **79.7** | 78.1 | 68.7 | 73.1 |
| Tree | 75.6 | 81.4 | 78.4 | 77.1 | 70.1 | 73.4 |
| Our model | **85.2** | 52.6 | 65.0 | 70.5 | 61.6 | 65.7 |
| Tree | 79.5 | 71.9 | 75.5 | 76.7 | **70.9** | **73.7** |

Table 6: Per-word labeling precision and recall of our model compared to the Tags-Small and Tags-Full baselines, both with and without expansion by trees. Our model is most precise on aspect and has better recall on value. Note that in general the process of expanding labels with the tree structure increases recall at the expense of precision.

**Results**   We evaluate all systems on precision and recall for aspect and value separately. Results for all systems are shown in Table 6. Our model without the tree expansion is highly precise at the expense of recall; however when the expansion is performed, its recall improves tremendously, especially on value words.

While this result is initially disappointing, it is possible to adjust model parameters to increase performance at this task; for example, for aspect words we could put additional

> The **moqueca** was **delicious** and **perfect** winter food , **warm** , filling and hearty but **not too heavy** .
> The **bacon** wrapped almond **dates** were **amazing** but the **plantains** with cheese were **boring** .
> the **artichoke** and homemade **pasta** appetizers were **great**

Table 7: High-precision, low-recall aspect word labeling by our full model. Note that a human would likely identify complete phrases such as *bacon wrapped almond dates* and *homemade pasta appetizers*; however, the additional noise degrades performance on the clustering task.

|        | A    | V    | B    | I    | end  |
|-------:|------|------|------|------|------|
| start  | 0.06 | 0.00 | 0.94 | 0.00 | 0.00 |
| A      | 0.19 | 0.03 | 0.77 | 0.01 | 0.00 |
| V      | 0.02 | 0.32 | 0.47 | 0.01 | 0.18 |
| B      | 0.22 | 0.26 | 0.43 | 0.17 | 0.06 |
| I      | 0.00 | 0.00 | 0.01 | 0.99 | 0.00 |

Table 8: Learned transition distribution from our model. The pattern of high-precision of aspect words is represented by a preference against continuing a string of several aspect words, causing the model to prefer single, precise aspect words. Likewise, the better recall of value words is indicated by a higher value of the `V V` transition, which can encourage several words in a row to be marked as value words.

mass on the prior for $Z_W^{i,j,w} = A$ or increase the Dirichlet hyperparameter $\lambda_A$. However, while this increases performance on the word labeling task, it also decreases performance correspondingly on the clustering task. By examination of the data, this correlation is perfectly reasonable. In order to succeed at the clustering task, the model selects only the most relevant portions of the snippet as aspect words. When the entire aspect and value are identified, clustering becomes noisy. Table 7 shows some examples of the high-precision labeling which achieves high clustering performance, and Table 8 shows an example of the learned transition distribution which creates this labeling.

## 6.2 Aspect Identification with Shared Aspects

Our second task uses a simplified version of our model designed for aspect identification only. For this task, we use a corpus of medical visit summaries. In this domain, each summary is expected to contain similar relevant information; therefore, the set of aspects is shared corpus-wide. To evaluate our model in this formulation, we examine the predicted clusters of snippets, as in the full model.

### 6.2.1 Data Set

Our data set for this task consists of phrases selected from dictated patient summaries at the Pediatric Environmental Health Clinic (PEHC) at Children's Hospital Boston, specializing in treatment of children with lead poisoning. Specifically, after a patient's office visit and lab results are completed, a PEHC doctor dictates a letter to the referring physician containing information about previous visits, current developmental and family status, in-office exam results, lab results, current diagnosis, and plan for the future.

For this experiment, we select phrases from the in-office exam and lab results sections of the summaries. Phrases are separated heuristically on commas and semicolons. In a domain which contains a significant amount of extraneous information, such as the restaurant domain, we must extract phrases which we believe bear some relevance to the task at hand. However, because the medical text is dense and nearly all relevant, a heuristic separation is sufficient to extract relevant phrases. There are 6198 snippets in total, taken from 271 summaries. The average snippet length is 4.5 words, and there are an average of 23 snippets per summary. As in the Yelp domain, we use the MXPOST tagger (Ratnaparkhi, 1996) to gain POS tags. Figure 12 shows some example snippets. For this domain, there are no values; we simply concentrate on the aspect-identification task. Unlike the restaurant domain, we use no seed words.

### 6.2.2 DOMAIN CHALLENGES AND MODELING TECHNIQUES

In contrast to the restaurant domain, the medical domain uses a single global set of aspects. These represent either individual lab tests (e.g., *lead level*, *white blood cell count*) or particular body systems (e.g., *lungs* or *cardiovascular*). Some aspects are far more common than others, and it is very uncommon for a summary to include more than one or two snippets about any given aspect. Therefore, as mentioned in Section 4.2, we model the aspect word distributions and the aspect multinomial as shared between all entities in the corpus.

Also in contrast to the restaurant domain, aspects are defined by words taken from the entire snippet. Rather than having aspects only associated with names of measurements (e.g., 'weight'), units and other descriptions of measurement (e.g., 'kilograms') are also relevant for aspect definition. This property extends to both numeric and written measurements; for example, the aspect 'lungs' is commonly described as 'clear to auscultation bilaterally'. In order to achieve high performance, our model must leverage all of these clues to provide proper aspect identification when the name of the measurement is missing (e.g., "patient is 100 cm"). While part of speech will still be an important factor to model, we predict that there will be greater importance on additional parts of speech other than nouns.

Finally, our data set is noisy and contains some irrelevant snippets, such as section headings (e.g., "Physical examination and review of systems") or extraneous information. As described in Section 4.2, we modify our model so that it can ignore partial or complete snippets.

### 6.2.3 CLUSTER PREDICTION

As for joint aspect and sentiment prediction, the goal of this task is to evaluate the quality of aspect identification. Because the aspects are shared across all documents, clusters are generally much larger, and the set of annotated snippets represents only a fraction of each cluster.

**Annotation**  For this experiment, we use a set of gold clusters gathered over 1,200 snippets, annotated by a doctor who is an expert in the domain from the Pediatric Environmental Health Clinic at Children's Hospital Boston. Note that as mentioned before, clusters

> He was 113 **cm** in **height**
> Patient's **height** was 146.5 **cm**

> **Lungs**: **Clear bilaterally** to **auscultation**
> **lungs** were normal

> **Heart** regular **rate** and **rhythm**; no **murmurs**
> **Heart** normal **S1 S2**

Figure 12: Example snippets from the medical data set, grouped according to aspect. Aspect words are underlined and colored blue. This grouping and labeling are *not* given in the data set and must be learned by the model.

|  | Precision | Recall | F1 |
|---|---|---|---|
| Cluster-All | 88.2 | 93.0 | 90.5 |
| Cluster-Noun | 88.4 | 83.9 | 86.1 |
| Our model | **89.1** | **93.4** | **91.2** |

Table 9: Results using the MUC metric on cluster prediction for the aspect identification only task. Note that the Cluster-All baseline significantly outperforms Cluster-Noun, the opposite of what we observe in the joint aspect and value prediction task. This is due to the dependence of aspect identification on more than just the name of a lab test, such as the units or other description of the test results, as mentioned in Section 6.2.2.

here are global to the domain (e.g., many patients have snippets representing *blood lead level*, and these are all grouped into one cluster). The doctor was asked to cluster 100 snippets at a time (spanning several patients), as clustering the entire set would have been infeasible for a human annotator. After all 12 sets of snippets were clustered, the resulting clusters were manually combined to match up similar clusters from each set. For example, the *blood lead level* cluster from the first set of 100 snippets was combined with the corresponding *blood lead level* clusters from each other set of snippets. Any cluster from this final set with fewer than 5 members was removed. In total, this yields a gold set of 30 clusters. There are 1,053 snippets total, for an average of 35.1 snippets per cluster. To match this, baseline systems and our full model are asked to produce 30 clusters across the full data set.

**Baselines & Metric** To keep these results consistent with those on the previous task, we use the same baselines and evaluation metric. Both baselines rely on a TF*IDF-weighted clustering algorithm, specifically implemented with CLUTO package (Karypis, 2002) using agglomerative clustering with the cosine similarity distance metric. As before, Cluster-All represents a baseline using unigrams of snippets from the entire data set, while Cluster-Noun works over only the nouns from the snippets. We again use the MUC cluster evaluation metric for this task. For more details on both baselines and the evaluation metric, please see Section 6.1.3.

**Results**  For this experiment, our system demonstrates an improvement of 7% over the CLUSTER-ALL baseline. Absolute performance is relatively high for all systems in the medical domain, indicating that the lexical clustering task is less misleading than in the restaurant domain. It is interesting to note that unlike in the restaurant domain, the CLUSTER-ALL baseline outperforms the CLUSTER-NOUN baseline. As mentioned in Section 6.2.2, the medical data is notable for the relevance of the entire snippet for clustering (e.g., both 'weight' and 'kilograms' are useful to identify the *weight* aspect). Because of this property, using only nouns to cluster in the CLUSTER-NOUN baseline hurts performance significantly.

## 7. Conclusions and Future Work

In this paper, we have presented an approach for fine-grained content aggregation using probabilistic topic modeling techniques to discover the structure of individual text snippets. Our model is able to successfully identify clusters of snippets in a data set which discuss the same aspect of an entity as well as the associated values (e.g., sentiment). It requires no annotation, other than a small list of seed vocabulary to bias the positive and negative distributions in the proper direction.

Our results demonstrate that delving into the structure of the snippet can assist in identifying key words which are important and unique to the domain at hand. When there are values to be learned, the joint identification of aspect and value can help to improve the quality of the results. The word labeling analysis reveals that the model learns a different type of labeling for each task; specifically, a strict, high-precision labeling for the clustering task and a high-recall labeling for sentiment. This follows the intuition that it is important to identify specific main points for clustering, while in the sentiment analysis task, there may often be several descriptions or conflicting opinions presented which all need to be weighed together to determine the overall sentiment.

This model admits a fast, parallelized inference procedure. Specifically, the entire inference procedure takes roughly 15 minutes to run on the restaurant corpus and less than 5 minutes on the medical corpus. Additionally, the model is neatly extensible and adjustable to fit the particular characteristics of a given domain.

There are a few limitations of this model which can be improved with future work: First, our model makes no attempt to explicitly model negation or other word interactions, increasing the difficulty of both aspect and sentiment analysis for our model. By performing error analysis, we find that negation is a common source of error for the sentiment analysis task. Likewise, on the aspect side, the model can make errors when attempting to differentiate aspects such as *ice cream* and *cream cheese* which share the common aspect word *cream*, despite these phrases occurring as bigrams. By using these connections in a stronger way, such as with an indicator variable for negation or a higher-order HMM, the model could make more informed decisions.

Second, while defining aspects per-entity as in the restaurant domain has advantages in that it is possible to get a very fine-grained set of applicable aspects, it also fails to leverage some potential information in the data set. Specifically, we know that restaurants sharing the same type (e.g., Italian, Indian, Bakery, etc.) should share some common aspects; however, there are no ties between them in the current model. Likewise, even at a global level, there may be some aspects which tie in across all restaurants. A hierarchical version of this model would be able to tie these together and identify different types of aspects: global (e.g., *presentation*), type-level (e.g., *pasta* for the Italian type), and restaurant-level (e.g., the restaurant's special dish).

## Bibliographic Note

Portions of this paper have been published previously in a conference publication (Sauper, Haghighi, & Barzilay, 2011); however this paper significantly extends that work. We describe several model generalizations and extensions (Section 4.2) and their effects on our inference procedure (Section 5.2). We present new experimental results, including additional baseline comparisons and an additional experiment (Section 6.1). We also introduce a new domain, medical summary text, which is quite different than the domain of restaurant reviews and therefore requires several fundamental changes to the model (Section 6.2).

## Acknowledgments

The authors acknowledge the support of the NSF (CAREER grant IIS-0448168), NIH (grant 5-R01-LM009723-02), Nokia, and the DARPA Machine Reading Program (AFRL prime contract no. FA8750-09-C-0172). Thanks to Peter Szolovits and the MIT NLP group for their helpful comments. Any opinions, findings, conclusions, or recommendations expressed in this paper are those of the authors, and do not necessarily reflect the views of the funding organizations.